\pdfoutput=1
\documentclass{article}

\usepackage[eandd, final]{neurips_2026}
\usepackage{amsmath}
\usepackage{enumitem}
\usepackage{graphicx}
\usepackage{placeins}
\usepackage[table]{xcolor}
\usepackage{longtable}
\usepackage{multirow}
\usepackage{float}

\usepackage[utf8]{inputenc} 
\usepackage[T1]{fontenc}    
\usepackage{hyperref}       
\usepackage{url}            
\usepackage{booktabs}       
\usepackage{amsfonts}       
\usepackage{nicefrac}       
\usepackage{microtype}      

\title{MOOD: Benchmarking Post-Hoc OOD Detection for Materials Property Prediction}

\author{%
  Shirley X.~Yu\textsuperscript{1,3} \qquad Amey P.~Pasarkar\textsuperscript{2,3} \qquad Adji Bousso Dieng\textsuperscript{1,3} \\[4pt]
  \textsuperscript{1}Department of Computer Science, Princeton University \\
  \textsuperscript{2}Lewis-Sigler Institute for Integrative Genomics, Princeton University \\
  \textsuperscript{3}Vertaix
}

\begin{document}

\maketitle

\begin{abstract}
Out-of-distribution (OOD) detection is essential for deploying machine learning models in high stakes scientific applications, yet methods that flag OOD inputs without retraining the model have been validated almost exclusively on image classification. We introduce MOOD (Materials Out-of-Distribution Detection), a benchmark for evaluating whether these post-hoc OOD detectors transfer to graph neural networks (GNNs) on materials property prediction. MOOD evaluates 15 post-hoc detectors and 2 supervised diagnostic probes across 4 GNN architectures, 5 MatBench tasks, and 15 physically motivated compositional and structural shifts, yielding 264 evaluation settings and over 13,000 AUROC measurements across three encoder training seeds. Our evaluation reveals three key findings. First, current post-hoc detectors perform near chance; all 15 detectors achieve an AUROC around 0.5. Second, detector failures decompose into two regimes: method bottlenecks and encoder bottlenecks. Method bottlenecks arise when supervised probes can separate in-distribution (ID) from OOD samples but unsupervised detectors fail, meaning that the embeddings contain OOD signal that current detectors do not extract. Encoder bottlenecks arise when even supervised probes fail to separate ID from OOD samples, indicating that the relevant shift is not accessible in the frozen GNN representation. Third, when OOD prediction error exceeds ID error, the AUROC of distance-based and density-based detectors correlates with the size of that gap. Collectively, these findings indicate that reliable OOD detection in scientific domains requires representation learning strategies that explicitly preserve OOD relevant signal, particularly for tasks where current GNN architectures encode no extractable signal at all.
\end{abstract}

\begin{center}
\small
Code: \url{https://github.com/vertaix/mood-benchmark} \quad 
\\
Data: \url{https://huggingface.co/datasets/shirlyu/mood-benchmark}
\end{center}

\section{Introduction}
\label{sec:introduction}

\looseness=-1 Machine learning (ML) models often encounter inputs that differ from their training data, making out-of-distribution (OOD) detection important for reliable deployment. Many OOD detection methods have been developed and benchmarked extensively in computer vision, especially for image classification. However, OOD detection remains far less explored in scientific ML, where data are often non-image, models are trained for regression, and distribution shifts reflect domain-specific structure.

\begin{figure}[t]
  \centering
  \includegraphics[width=\linewidth]{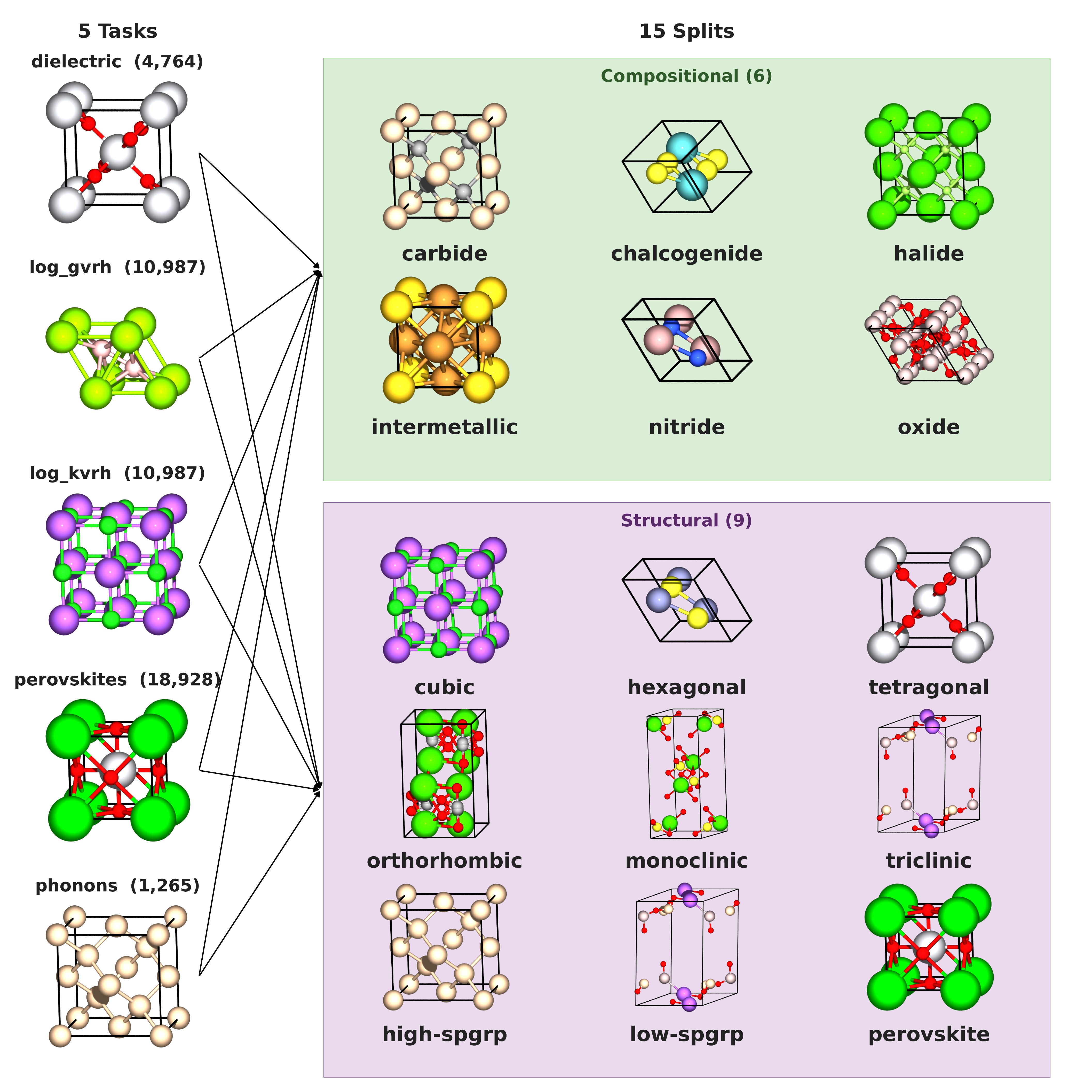}
 \caption{\textbf{MatBench tasks with sample sizes and OOD splits applied to each.} Every panel shows a representative material. On the left is a general example from the task's dataset: rutile TiO$_2$ for dielectric, MgB$_2$ for log\_gvrh, NaCl for log\_kvrh, SrTiO$_3$ for perovskites, diamond Si for phonons. On the right is an example belonging to the split's OOD class: SiC for carbide, NaCl for cubic, SrTiO$_3$ for perovskite, etc.}
  \label{fig:datasets_splits}
\end{figure}

Addressing this gap is especially important for scientific ML because model predictions often guide expensive downstream simulations or experiments. When deployment inputs differ from the training distribution, unreliable predictions can waste substantial computational, experimental, or synthesis effort. In materials discovery, this risk is built in. Screening pipelines deliberately
seek candidates with new compositions or structures, which are exactly the inputs
where graph neural network (GNN) predictions degrade~\citep{omee2024structurebased}, and a regression
model gives little to no signal that it is extrapolating. This setting stresses OOD detection differently from computer vision because GNNs for materials property prediction are typically trained for regression, so their embeddings are not organized around class boundaries in the way image classifier embeddings often are. Recent work has benchmarked predictive performance under distribution shift for GNNs~\citep{gui2022good, ji2023drugood} and materials property prediction models~\citep{omee2024structurebased, tan2025matuq}, but OOD detection for frozen pre-trained materials GNNs has not been systematically benchmarked.

As such, we introduce MOOD (Materials Out-of-Distribution Detection), a benchmark for evaluating OOD detection on frozen embeddings from GNNs trained for materials property prediction (Figure~\ref{fig:datasets_splits}). This setting is useful for materials discovery pipelines because detectors can be added after a GNN property predictor is trained, without modifying the model or rerunning the training pipeline~\citep{sun2022knn, lee2018mahalanobis, djurisic2023ash, wang2022vim}. Following prior work, we call these detectors ``post-hoc,'' meaning they score frozen model embeddings at inference time.

Our evaluation reveals three findings. First, all 15 unsupervised post-hoc detectors perform near chance, with the strongest reaching only $0.531$ mean AUROC, indicating that the vision OOD post-hoc toolkit does not transfer. Second, this near chance performance reflects two distinct failure regimes: \textbf{method bottlenecks} and \textbf{encoder bottlenecks}. In method bottlenecks, supervised probes can separate ID from OOD samples, but unsupervised detectors fail to do so, indicating that the embeddings contain OOD signal that current detectors do not extract. In encoder bottlenecks, both probes and detectors fail to separate ID from OOD samples, indicating that the relevant shift is not encoded in the frozen GNN representation. Third, when the predictor performs worse on OOD samples than on ID samples, distance-based and density-based detector AUROC correlates with the OOD-to-ID error gap. These regimes point to different interventions. In method-bottlenecked settings, better detectors are needed to extract OOD signal already present in the embeddings; in encoder-bottlenecked settings, better encoders are needed to preserve OOD relevant signal.

\paragraph{Contributions.} We make three main contributions.
\begin{itemize}[leftmargin=*,itemsep=2pt,topsep=2pt]
\item We introduce \textbf{MOOD}, the first systematic benchmark for post-hoc OOD detection on GNNs trained for materials property prediction. It covers 264 evaluation settings across 5 tasks, 4 architectures, and 15 physically motivated shifts. We release the embeddings, data splits, and evaluation code.
\item We propose a \textbf{diagnostic decomposition} that sorts detector failures into two types. In a method bottleneck, the frozen embeddings contain ID/OOD signal, but current detectors fail to extract it. In an encoder bottleneck, the embeddings contain no ID/OOD signal that a probe can decode. We tell the two apart with supervised probes, which measure how much OOD signal a classifier can decode from each embedding.
\item We provide an \textbf{empirical characterization} of when post-hoc OOD detection works, based on over 13,000 AUROC measurements. All detectors perform near chance overall. However, distance- and density-based detectors improve when the GNN predicts OOD samples worse than ID samples, and their AUROC tracks the size of this error gap.
\end{itemize}

\section{Related Work}
\label{sec:related}

\paragraph{Post-hoc OOD detection on frozen embeddings.}
Post-hoc detectors score samples using a pre-trained model's penultimate layer embeddings without retraining. Distance-based methods score proximity to the training data, including $k$-nearest-neighbor (KNN) distance \citep{sun2022knn}, embedding norm \citep{yu2023featurenorm}, and cosine similarity to the nearest training neighbor \citep{tack2020csi, sun2022knn}. Density-based methods estimate the training distribution explicitly, including parametric Mahalanobis models with class conditional Gaussian components \citep{lee2018mahalanobis}, kernel density estimation \citep{bishop1994novelty}, and normalizing flows \citep{dinh2017realnvp}. Classical anomaly-detection methods such as Isolation Forest \citep{liu2008iforest}, one class SVM \citep{scholkopf1999ocsvm}, and local outlier factor \citep{breunig2000lof} provide additional baselines. These methods are appealing for deployment because they operate at inference time on a fixed pre-trained model, without retraining. This matches materials screening pipelines, where a trained GNN predicts properties for large pools of candidate materials.

\paragraph{Post-hoc embedding space detection.}
MOOD focuses on detectors that score frozen embeddings from a
pre-trained GNN. This setting matches the dominant post-hoc OOD
detection literature and reflects a common scientific ML workflow, where
a trained encoder is reused across downstream property prediction tasks
without modification. This scope separates MOOD from three related settings. Training time OOD
methods \citep{hendrycks2019oe, liu2020energy, winkens2020contrastive,
katz2022training} modify the training objective or require retraining
the model, often with auxiliary outlier data. Confidence-based and
internal layer methods
\citep{hendrycks2017baseline, lakshminarayanan2017deepensembles,
gal2016dropout, spectralgap2024} rely on predictive uncertainty,
ensembles, stochastic inference, or architecture specific activations.
Domain adaptation methods \citep{wilson2020dasurvey} assume access to
target domain data and aim to improve model performance under shift.
MOOD instead isolates a complementary question: how much OOD signal can be recovered from frozen GNN embeddings?

\paragraph{OOD benchmarks.}
The Open Graph Benchmark \citep{hu2020ogb} established standardized dataset splits for evaluating graph ML models, which later graph benchmarks build on. OpenOOD \citep{yang2022openood} provides one of the most comprehensive vision OOD detection benchmarks, evaluating 30+ methods across image classification tasks. GOOD \citep{gui2022good} and DrugOOD \citep{ji2023drugood} benchmark OOD generalization on graphs and molecules respectively, and BOOM \citep{antoniuk2025boom} benchmarks extrapolation to out-of-range molecular property values; all three evaluate model robustness rather than post-hoc detection. GeSS \citep{zou2024gess} benchmarks geometric deep learning under distribution shift across multiple scientific domains with three levels of OOD information access, focusing on prediction accuracy rather than detection. In materials specifically, the structure-based OOD benchmark of \citet{omee2024structurebased} evaluates property prediction performance using five clustering-based splits on three MatBench datasets, and MatUQ \citep{tan2025matuq} benchmarks uncertainty quantification across six materials datasets under structural distribution shifts. Both focus on prediction quality and uncertainty calibration rather than identifying which test samples are OOD in the first place. BROAD \citep{guilleescuret2024broad} is the closest methodological precedent, because it benchmarks post-hoc OOD detection in ImageNet vision classification across five shift types. MOOD asks the analogous question for materials GNN property prediction, using 15 physically motivated compositional and structural shifts, and introduces supervised diagnostic probes to decompose detector failures into method and encoder bottlenecks.

\paragraph{OOD generalization on materials GNNs.}
Materials property prediction GNNs achieve strong in-distribution (ID)
performance on MatBench \citep{dunn2020matbench, choudhary2021alignn},
but their OOD generalization is fragile~\citep{omee2024structurebased}. \citet{omee2024structurebased} show
that state-of-the-art GNNs significantly underperform their
MatBench baselines on structurally OOD test sets, with mean absolute
error (MAE) degradation reaching $5\times$ on some splits.
\citet{tan2025matuq} find that OOD error persists even under uncertainty-aware training. These findings motivate the central question
MOOD addresses: can post-hoc methods reliably flag unreliable GNN predictions on OOD inputs?

\FloatBarrier

\section{Benchmark Design}
\label{sec:design}

We construct a benchmark for evaluating post-hoc OOD detection on frozen embeddings from materials property prediction GNNs. Each evaluation setting combines one MatBench task, one GNN architecture, and one OOD split, where a split is a rule that partitions a dataset into ID and OOD samples (Figure~\ref{fig:datasets_splits}). The 5 tasks, 4 architectures, and 15 OOD splits yield 300 possible settings, of which 264 are retained after filtering small OOD groups. Seven CGCNN/phonons encoders collapse to constant embeddings under seed 42 only and train normally under seeds 123 and 456; for these settings we average over the two valid seeds. The full benchmark schematic is in Appendix~\ref{app:dataset_details}.

\subsection{Tasks and architectures}
We use 5 regression tasks from MatBench~\citep{dunn2020matbench}, a standard materials property prediction dataset suite, with each target shown in parentheses:
dielectric (refractive index), log\_gvrh (log shear modulus), log\_kvrh
(log bulk modulus), perovskites (formation energy), and phonons (phonon
frequency). These span optical, mechanical, thermodynamic, and
vibrational properties. We evaluate four GNN architectures:
ALIGNN~\citep{choudhary2021alignn} ($d=256$),
CGCNN~\citep{xie2018cgcnn} ($d=128$), SchNet~\citep{schutt2017schnet}
($d=64$), and DimeNet++~\citep{gasteiger2020dimenet} ($d=64$). Each is trained with three random seeds (42, 123, 456) on a 90/10 random split of the ID set, using default hyperparameters from the original implementations; we report mean $\pm$ std across seeds. Embeddings are
extracted as penultimate layer activations and held fixed during
detection, making the comparison strictly post-hoc.

\subsection{OOD splits}
We construct 15 deterministic OOD splits, 6 compositional and 9 structural. The compositional splits group materials by what elements they contain: carbide, chalcogenide, halide, intermetallic, nitride, oxide. The structural splits group materials by crystal shape and symmetry, and break down into 6 crystal shape categories (cubic, hexagonal, tetragonal, orthorhombic, monoclinic, triclinic), 2 symmetry levels (high and low), and 1 group of perovskite-shaped materials. Each split is a binary rule that partitions every material into one of two groups based on whether it has the compositional or structural property in question. Full definitions are in Appendix~\ref{app:dataset_details}.

For each split, the two groups are sorted by size. The majority becomes the ID pool and the minority becomes the OOD test set, so group size---not which side of the rule a material falls on---determines the ID/OOD assignment. The ID pool is partitioned 90/10 into train and test with a fixed random seed. To enable class balanced AUROC, both test sets are subsampled to the same size, $\min(|\text{ID}_{\text{test}}|, |\text{OOD}_{\text{test}}|)$, so detectors are evaluated on equal numbers of ID and OOD samples.

\subsection{Detection methods}
We evaluate 15 unsupervised detectors in four mechanism families, plus two supervised diagnostic probes. \textbf{Distance (6):} KNN $k{=}20$~\citep{sun2022knn}, KNN $k{=}50$, embedding norm~\citep{yu2023featurenorm}, Cosine~\citep{tack2020csi, sun2022knn}, LOF~\citep{breunig2000lof}, ASH+KNN~\citep{djurisic2023ash}. \textbf{Density (5):} MDS-BIC~\citep{lee2018mahalanobis}, Gaussian KDE~\citep{bishop1994novelty}, Real-NVP normalizing flow~\citep{dinh2017realnvp}, kernel PCA reconstruction error~\citep{hoffmann2007kpca}, ReAct+MDS~\citep{sun2021react}. \textbf{Isolation (2):} Isolation Forest~\citep{liu2008iforest}, ADERH~\citep{yang2024aderh}. \textbf{Boundary/Hybrid (2):} one-class SVM~\citep{scholkopf1999ocsvm}, ViM~\citep{wang2022vim} applied as residual norm only since our regression GNNs lack a class manifold for the original logit-space component. \textbf{Supervised diagnostic probes (2):} a linear probe and an MLP probe, trained on ID/OOD labels under 5 fold stratified cross validation. Probes are diagnostic ceilings, not deployable detectors, because they require OOD labels at fit time and bound the OOD signal that a linear or MLP classifier can decode from the frozen embedding, not all information the embedding may contain. Full hyperparameters and detector implementations are in Appendix~\ref{app:detector_details}.

\subsection{Bottleneck classification}
Detector performance is measured by AUROC on balanced ID/OOD test sets. The family-max is the best AUROC achieved within each family (Distance, Density, Isolation, Boundary/Hybrid) per setting. To establish whether probe performance reflects genuine ID/OOD separability rather than overfitting, we run a per-setting permutation test with $B = 100$ label shuffles under the same 5 fold protocol, computing $p = (1 + \#\{\text{null} \geq \text{observed}\}) / (B + 1)$ and $z = (\text{observed} - \mu_{\text{null}}) / \sigma_{\text{null}}$. Under the null, $\sim$1\% of settings should reach $p < 0.01$ by chance. Results are in Appendix~\ref{app:numerical_results}, Table~\ref{tab:permutation_test}.

\looseness=-1
We classify each task as method-bottlenecked, encoder-bottlenecked, or mixed, where probe gap is the mean linear probe AUROC minus the mean AUROC of the 15 unsupervised detectors. A task is method-bottlenecked if linear probe AUROC $\geq 0.55$ and probe gap $\geq 0.05$, meaning signal exists in the embeddings but detectors fail to extract it. It is encoder-bottlenecked if probe AUROC $< 0.55$, meaning no signal that a linear or MLP probe can decode exists in the embeddings. It is mixed if probe AUROC $\geq 0.55$ and probe gap $< 0.05$, meaning both probe and detector recover signal, or both sit just above chance. The 0.55 probe threshold and 0.05 probe gap threshold are conservative cutoffs; we discuss their justification in Appendix~\ref{app:author_statement}. Our primary probe gap is measured against the mean of all 15 detectors. We also report it against the single best detector (Cosine). Under this stricter definition, both method-bottlenecked tasks fall below the 0.05 gap cutoff and would be labeled Mixed, while the encoder-bottleneck labels, which depend on probe AUROC alone, are unchanged.

\subsection{Linking detection to regression error}
To test whether detection performance tracks distribution shift magnitude, we define the per-setting MAE gap as $\log_{10}(\text{MAE}_{\text{OOD}} / \text{MAE}_{\text{ID}})$, where MAE is the GNN's regression error. For each detector we fit an Ordinary Least Squares (OLS) regression
\begin{equation}
\text{AUROC}_s = \alpha + \beta \cdot \log_{10}\frac{\text{MAE}_{\text{OOD},s}}{\text{MAE}_{\text{ID},s}} + \varepsilon_s
\end{equation}
reported on all $N_{all} = 264$ settings and on the degraded subset $N_{deg} = 153$ (settings where $\mathrm{MAE}_{\mathrm{OOD}} > \mathrm{MAE}_{\mathrm{ID}}$ with both MAEs averaged over the three encoder seeds, the regime in which the GNN actually degrades on OOD). The MAE gap is a retrospective mechanism check on when distance- and density-based detectors have purchase; it requires OOD labels to compute and cannot be used to decide in advance whether detection will work.

\section{Evaluation}
\label{sec:evaluation}

\begin{table}[t]
\centering
\small
\setlength{\tabcolsep}{3pt}
\caption{\textbf{Per-detector performance summary across $N_{all}=264$ settings} (mean $\pm$ std over three encoder training seeds). Win count splits ties equally; \% settings $>0.55$ uses the above chance threshold of \S\ref{sec:design}. Supervised probes (bottom two rows) are diagnostic ceilings, not deployable detectors. \textbf{Bold} is the best unsupervised detector per family. $\uparrow$ indicates higher is better.}
\label{tab:per_detector_summary}
\resizebox{\linewidth}{!}{%
\begin{tabular}{llcccc}
\toprule
Family & Detector & Mean AUROC $\uparrow$ & Median AUROC $\uparrow$ & Win count $\uparrow$ & \% settings $>0.55$ $\uparrow$ \\
\midrule
\multirow{6}{*}{Distance} & KNN-k20-norm & 0.512 $\pm$ 0.001 & 0.504 $\pm$ 0.001 & 13.2 $\pm$ 3.0 & 10.4 $\pm$ 0.8\% \\
 & KNN-k50-norm & 0.508 $\pm$ 0.001 & 0.503 $\pm$ 0.001 & 7.7 $\pm$ 3.5 & 9.3 $\pm$ 1.8\% \\
 & Norm & 0.500 $\pm$ 0.001 & 0.500 $\pm$ 0.001 & 23.3 $\pm$ 4.0 & 6.1 $\pm$ 1.1\% \\
 & Cosine & \textbf{0.531 $\pm$ 0.001} & \textbf{0.507 $\pm$ 0.001} & \textbf{54.8 $\pm$ 8.6} & \textbf{13.5 $\pm$ 0.6\%} \\
 & LOF & 0.526 $\pm$ 0.001 & 0.505 $\pm$ 0.001 & 30.2 $\pm$ 3.3 & 13.1 $\pm$ 0.2\% \\
 & ASH+KNN & 0.516 $\pm$ 0.002 & 0.505 $\pm$ 0.002 & 30.8 $\pm$ 5.3 & 11.3 $\pm$ 0.6\% \\
\cmidrule(lr){1-6}
\multirow{5}{*}{Density} & MDS-BIC & 0.513 $\pm$ 0.001 & \textbf{0.504 $\pm$ 0.000} & 3.7 $\pm$ 2.3 & 8.5 $\pm$ 1.2\% \\
 & KDE & 0.518 $\pm$ 0.001 & \textbf{0.504 $\pm$ 0.001} & 16.7 $\pm$ 3.8 & \textbf{10.2 $\pm$ 0.4\%} \\
 & Flow & \textbf{0.520 $\pm$ 0.003} & 0.503 $\pm$ 0.001 & \textbf{17.3 $\pm$ 5.9} & 9.0 $\pm$ 0.1\% \\
 & kPCA & 0.512 $\pm$ 0.000 & 0.503 $\pm$ 0.000 & 5.7 $\pm$ 1.5 & 9.0 $\pm$ 1.4\% \\
 & ReAct+MDS & 0.512 $\pm$ 0.001 & 0.503 $\pm$ 0.000 & 6.0 $\pm$ 3.0 & 8.0 $\pm$ 1.3\% \\
\cmidrule(lr){1-6}
\multirow{2}{*}{Isolation} & IForest & \textbf{0.502 $\pm$ 0.001} & \textbf{0.500 $\pm$ 0.001} & \textbf{11.7 $\pm$ 4.5} & \textbf{7.4 $\pm$ 0.6\%} \\
 & ADERH & 0.497 $\pm$ 0.002 & \textbf{0.500 $\pm$ 0.001} & 11.2 $\pm$ 2.3 & 4.7 $\pm$ 0.9\% \\
\cmidrule(lr){1-6}
\multirow{2}{*}{\shortstack{Boundary/\\Hybrid}} & OCSVM & 0.496 $\pm$ 0.001 & 0.498 $\pm$ 0.001 & \textbf{19.3 $\pm$ 6.1} & 7.4 $\pm$ 1.6\% \\
 & ViM & \textbf{0.509 $\pm$ 0.001} & \textbf{0.503 $\pm$ 0.001} & 10.2 $\pm$ 2.0 & \textbf{8.5 $\pm$ 1.8\%} \\
\midrule
\multirow{2}{*}{Supervised} & Linear probe & \textbf{0.543 $\pm$ 0.001} & \textbf{0.500 $\pm$ 0.001} & -- & \textbf{16.4 $\pm$ 1.2\%} \\
 & MLP probe & 0.542 $\pm$ 0.002 & 0.496 $\pm$ 0.002 & -- & 15.5 $\pm$ 0.8\% \\
\bottomrule
\end{tabular}}
\end{table}

\begin{figure}[t]
  \centering
  \includegraphics[width=0.8\linewidth]{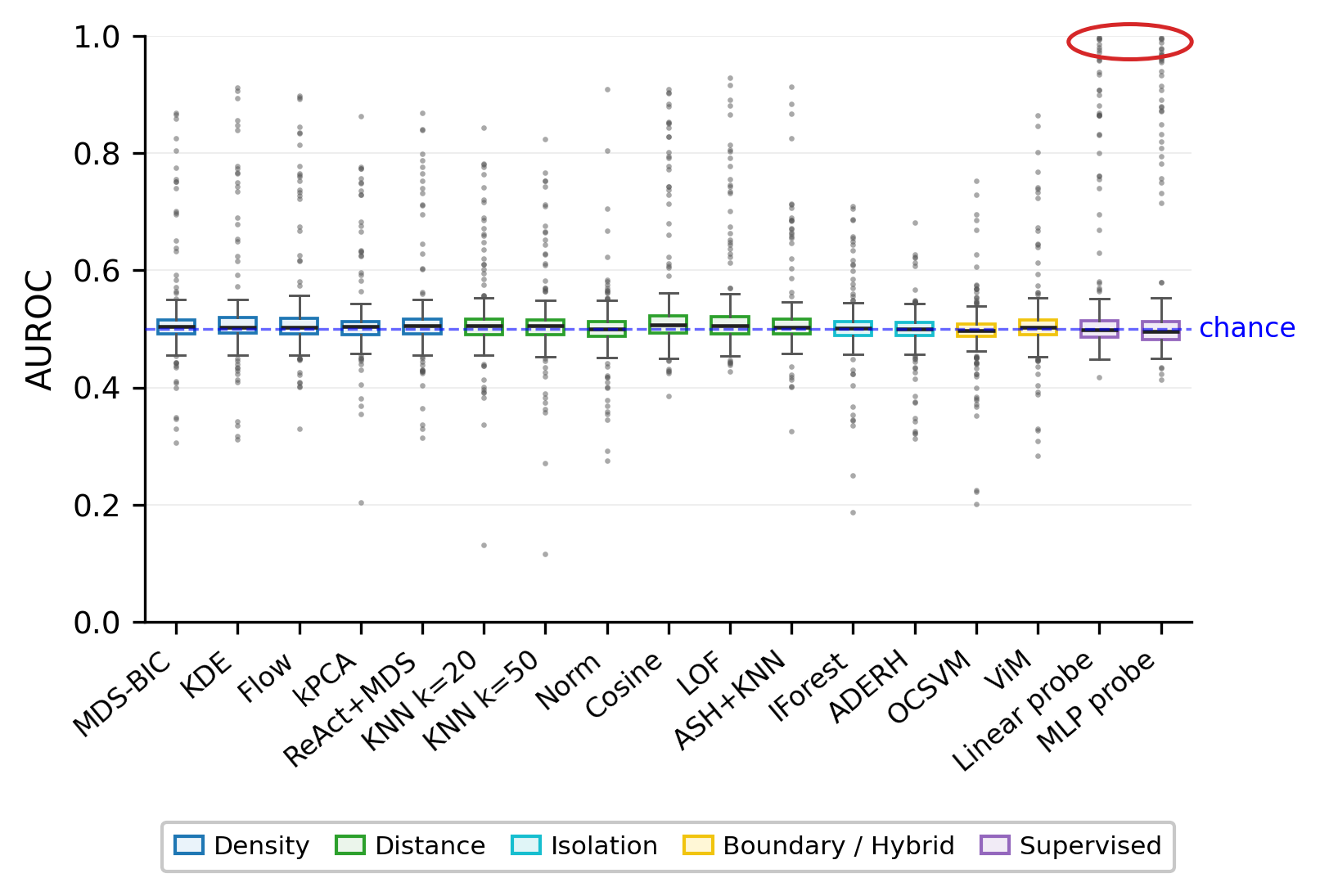}
  \caption{\textbf{Headline AUROC across $N_{all} = 264$ settings.} Every unsupervised detector median is within $0.01$ of chance. The
  supervised probes extend to AUROC $\approx 1.0$
  on multiple settings (circled in red), indicating extractable OOD signal exists in
  some embeddings.}
  \label{fig:headline_auroc}
\end{figure}

We organize our analysis around three findings: aggregate near chance performance (\S\ref{sec:eval_uniform}), bottleneck dichotomy (\S\ref{sec:eval_dichotomy}), and AUROC tracking the MAE gap (\S\ref{sec:eval_mae}); robustness analyses are in Appendix~\ref{app:robustness}.

\subsection{Post-hoc detectors fail uniformly across the benchmark}
\label{sec:eval_uniform}
Across all $N_{all} = 264$ settings, every unsupervised detector achieves a median AUROC within $0.01$ of chance (Table~\ref{tab:per_detector_summary}, Figure~\ref{fig:headline_auroc}). The strongest detector, Cosine, reaches mean AUROC $0.531$ and exceeds $0.55$ on only $13.5\%$ of settings; the weakest, ADERH, reaches mean $0.497$ with $4.7\%$ of settings above $0.55$. The supervised probes recover more signal (mean $0.543$ linear, $0.542$ MLP; $16.4\%$ and $15.5\%$ above $0.55$), but their medians remain near chance, so strong probe results come from a minority of settings.

\begin{figure}[t]
  \centering
  \includegraphics[width=\linewidth]{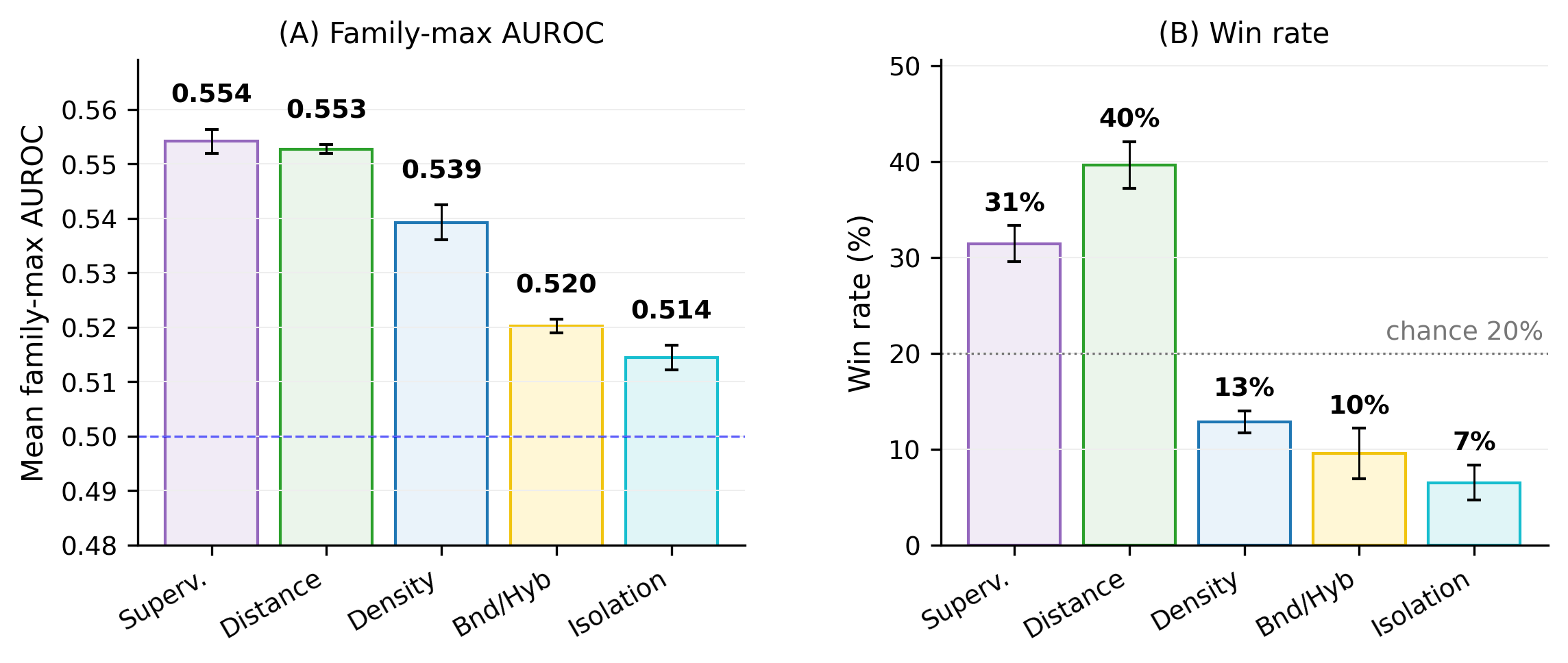}
  \caption{\textbf{Family level analysis.} (A) Per-setting family-max
  AUROC. Error bars are ±1 Standard Error of the Mean (SEM) across $N_{all}=264$ settings. (B) Win rate per family, including the supervised probes as a fifth group, with ties split equally.}
  \label{fig:family_analysis}
\end{figure}
Figure~\ref{fig:family_analysis} reports family-max AUROC, which is the best AUROC of all detectors within that family for each setting. Even under this best case per-setting view, every family sits only modestly above chance and below the supervised AUROC at $0.554$. Distance methods are the strongest of the four detector families, winning $40\%$ of settings against a $20\%$ random baseline over the five groups in Figure~\ref{fig:family_analysis}B (the four detector families plus the supervised probes); even so, their mean family-max AUROC of $0.553$ remains far below practical deployment thresholds. 

Detector scores are not entirely uninformative. In a risk coverage analysis on the seed 42 encoders, abstaining on the samples a detector flags as most anomalous removes high error predictions somewhat more often than random abstention. However, mean selective prediction effectiveness (SPE) ranges only from $0.14$ to $0.37$, where an oracle that abstains on the worst predictions first scores $1$ (Appendix~\ref{app:risk_coverage}). Current detectors therefore close at most $37\%$ of the gap between random and oracle abstention, leaving most unreliable predictions unflagged. The near chance ceiling is also not an artifact of our split construction. Under the three alternative split protocols of \citet{omee2024structurebased} (LOCO, SparseXcluster, and SparseYcluster), Mahalanobis and KNN detectors stay within about $0.05$ of chance on dielectric and log\_gvrh (Appendix~\ref{app:omee}).

\subsection{Failures decompose into method and encoder bottlenecks}
\label{sec:eval_dichotomy}
\begin{table}[t]
\centering
\small
\setlength{\tabcolsep}{4pt}
\caption{\textbf{Bottleneck dichotomy by task} (mean $\pm$ std over three encoder training seeds). Linear/MLP are the supervised probe AUROCs. Mean/Best detector are the mean across all 15 unsupervised detectors and Cosine (highest-mean detector). Gap (mean) $=$ Linear $-$ Mean detector (sets the Bottleneck label); Gap (best) $=$ Linear $-$ Best detector. On both method-bottlenecked tasks the probe exceeds even the best detector, by less than 0.05 (\S3.4, Appendix~H). Method columns shaded yellow, Encoder columns blue. \textbf{Bold} is the best value per row. $\uparrow$ indicates higher is better.}
\label{tab:bottleneck_summary}
\begin{tabular}{l >{\columncolor{yellow!20}}r >{\columncolor{yellow!20}}r >{\columncolor{blue!12}}r >{\columncolor{blue!12}}r >{\columncolor{blue!12}}r}
\toprule
& dielectric & log\_gvrh & log\_kvrh & perovskites & phonons \\
\midrule
$N$ & 56 & 60 & 60 & 36 & 52 \\
Linear $\uparrow$ & 0.599 $\pm$ 0.001 & \textbf{0.600 $\pm$ 0.002} & 0.494 $\pm$ 0.004 & 0.500 $\pm$ 0.005 & 0.499 $\pm$ 0.009 \\
MLP $\uparrow$ & 0.601 $\pm$ 0.005 & \textbf{0.606 $\pm$ 0.003} & 0.490 $\pm$ 0.003 & 0.499 $\pm$ 0.002 & 0.493 $\pm$ 0.008 \\
Mean det.~$\uparrow$ & 0.522 $\pm$ 0.002 & \textbf{0.524 $\pm$ 0.001} & 0.504 $\pm$ 0.001 & 0.500 $\pm$ 0.001 & 0.502 $\pm$ 0.003 \\
Best det.~$\uparrow$ & \textbf{0.569 $\pm$ 0.003} & 0.560 $\pm$ 0.001 & 0.505 $\pm$ 0.001 & 0.502 $\pm$ 0.001 & 0.506 $\pm$ 0.003 \\
Gap (mean) $\uparrow$ & \textbf{+0.078 $\pm$ 0.002} & +0.076 $\pm$ 0.002 & $-$0.010 $\pm$ 0.004 & +0.001 $\pm$ 0.005 & $-$0.003 $\pm$ 0.010 \\
Gap (best) $\uparrow$ & +0.031 $\pm$ 0.004 & \textbf{+0.040 $\pm$ 0.003} & $-$0.011 $\pm$ 0.003 & $-$0.002 $\pm$ 0.005 & $-$0.007 $\pm$ 0.010 \\
Bottleneck & Method & Method & Encoder & Encoder & Encoder \\
\bottomrule
\end{tabular}
\end{table}

The aggregate picture hides substantial variation across tasks, which fall into two distinct regimes. We classify each task by its bottleneck regime per \S\ref{sec:design}; Table~\ref{tab:bottleneck_summary} reports the assignment. Under the mean-detector probe gap, the five tasks partition into the two regimes, with none Mixed; against the single best detector (Cosine), dielectric and log\_gvrh would be Mixed, as the probe still exceeds Cosine but by less than $0.05$. Dielectric (probe $0.599$, gap $+0.078$) and log\_gvrh (probe $0.600$, gap $+0.076$) are method-bottlenecked. Log\_kvrh (probe $0.494$), perovskites (probe $0.500$), and phonons (probe $0.499$) are encoder-bottlenecked.

\begin{figure}[t]
  \centering
  \includegraphics[width=\linewidth]{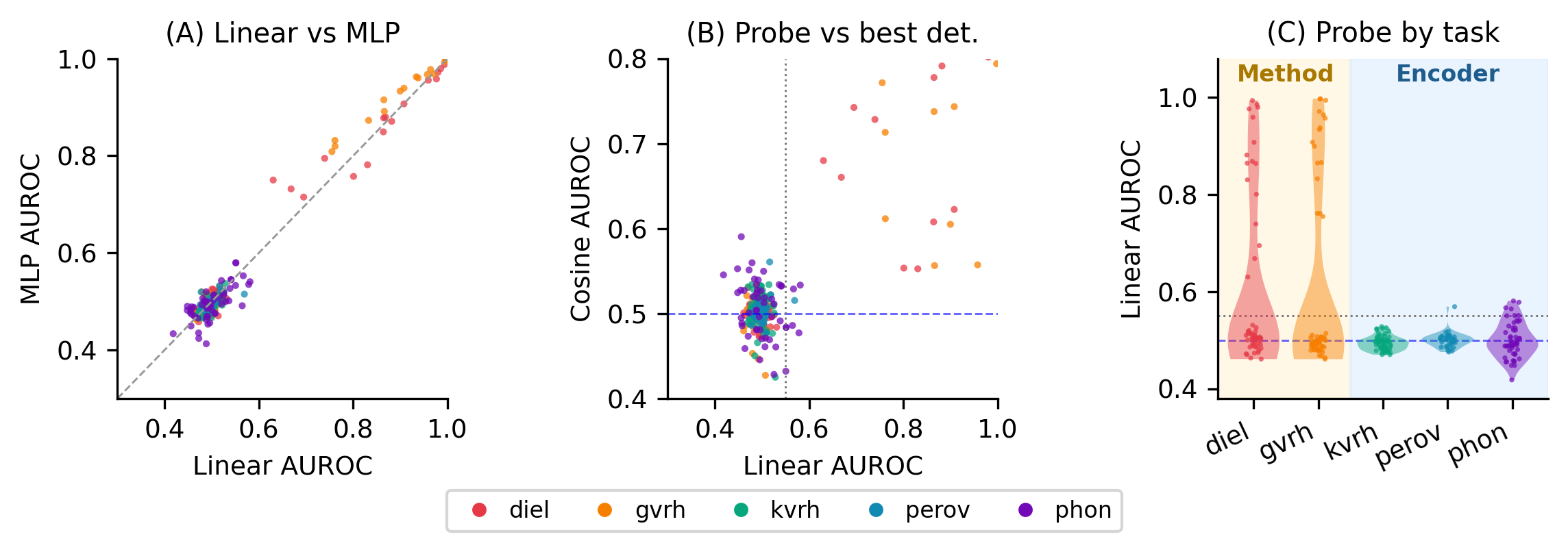}
    \caption{\textbf{Method vs.\ encoder bottlenecks.} (A) Linear vs.\ MLP probe AUROC; close agreement ($r{=}0.971$) rules out nonlinearity as the explanation for the encoder bottleneck. (B) Linear probe vs.\ Cosine (best deployable detector); two clusters emerge separating method-bottlenecked tasks (probe high, detector low) from encoder-bottlenecked ones (both near chance). (C) Linear probe AUROC by task; yellow (\textit{Method}) and blue (\textit{Encoder}) shading mark the two bottleneck regimes.}
  \label{fig:bottleneck_dichotomy}
\end{figure}

Figure~\ref{fig:bottleneck_dichotomy} illustrates the two regimes. Panel~(B) splits into two clusters. In the lower right group (method bottleneck), the probe reaches $0.7$--$1.0$ while Cosine remains near $0.5$--$0.6$, showing that signal is in the embeddings but unsupervised detectors cannot extract it. In the lower left group (encoder bottleneck), points sit near $(0.5, 0.5)$, showing that neither probes nor detectors can find OOD signal. Panel~(C) shows the same split by task. Probe AUROCs on dielectric and log\_gvrh spread from chance up to $1.0$, while those on the other three tasks stay at chance regardless of split. Panel~(A) rules out probe non-linearity as the explanation for the encoder-bottleneck regime, since the MLP probe agrees with the linear probe to within mean $|\Delta| = 0.023$ ($r = 0.971$) and never exceeds AUROC $0.629$ on encoder-bottlenecked settings.

To certify the dichotomy reflects genuine ID/OOD separability rather than overfitting, we run a per-setting permutation test on the linear probe (\S\ref{sec:design}); Table~\ref{tab:permutation_test} (Appendix~\ref{app:numerical_results}) reports rejection rates. The method-bottlenecked tasks reject the null at $p < 0.01$ in $28.6\%$ of settings (dielectric, $16/56$ under every seed) and $26.7\%$ (log\_gvrh, $16/60$ every seed), with mean $z$-scores of $+3.10$ and $+4.95$ and the strongest settings reaching $z = 18$--$33$. The encoder-bottlenecked tasks reject at $p < 0.01$ in at most $1$ of $52$ settings under any seed (phonons, seed 456), consistent with the $\sim$1\% expected under the null, and at $p < 0.05$ in $2$--$4\%$ of settings. Method-bottlenecked tasks exceed the chance floor by an order of magnitude or more; encoder-bottlenecked tasks sit at or below it. The dichotomy is statistically real and invariant to architecture; all 20 (dataset, architecture) cells fall on the same side of the 0.55 threshold as the dataset-level label (Table~\ref{tab:per_arch_probe}).

\paragraph{Stability across training seeds.} Across the three training seeds, the two regimes are separated by
$0.101$--$0.105$ in mean probe AUROC, while seed-to-seed standard deviations are $0.001$--$0.009$,
roughly an order of magnitude smaller than the gap a seed would need to close to change
a label. No dataset changes label under any seed or architecture. A minority of
individual settings cross the 0.55 cutoff on a single seed (Appendix~\ref{app:robustness}),
concentrated in phonons, the smallest task.

\subsection{Detector AUROC tracks regression-error degradation}
\label{sec:eval_mae}
\begin{figure}[t]
  \centering
  \includegraphics[width=\linewidth]{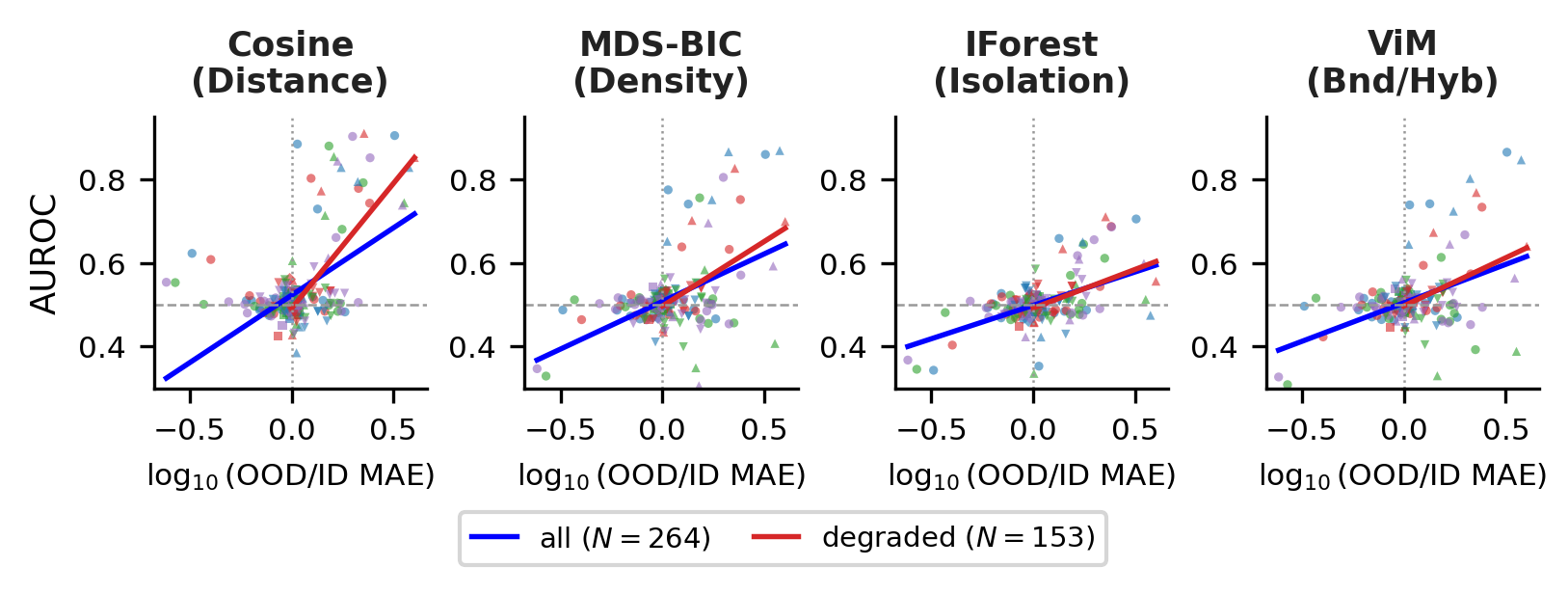}
  \caption{\textbf{Detector AUROC tracks regression failure magnitude.} Per-setting AUROC against $\log_{10}(\mathrm{MAE}_{\mathrm{OOD}} / \mathrm{MAE}_{\mathrm{ID}})$ for one representative detector per family. Blue lines fit $N_{all} = 264$ settings; red lines fit the degraded subset $N_{deg} = 153$. Per-detector grid in Appendix~\ref{app:numerical_results}, Figure~\ref{fig:auroc_vs_mae_gap_full}.}
  \label{fig:auroc_vs_mae_gap_compacted}
\end{figure}
\begin{table}[h]
\centering
\caption{\textbf{Regression of detector AUROC against $\boldsymbol{\log_{10}(\mathrm{MAE}_{\mathrm{OOD}}/\mathrm{MAE}_{\mathrm{ID}})}$.} OLS slope $\beta$ and $R^2$ for one representative detector per family plus the supervised probes, fit on the seed-averaged per-setting points (AUROC and MAE ratio averaged over the three encoder seeds) for all $N_{all}=264$ settings and the degraded subset $N_{deg}=153$. Per-seed fits ($\beta\pm$std) are in Appendix Table~\ref{tab:regression_auroc_mae_gap}. \textbf{Bold} is the best value per column. $\uparrow$ indicates higher is better.}
\label{tab:regression_compacted}
\small
\begin{tabular}{lrrrr}
\toprule
Detector & $\beta$ (all) $\uparrow$ & $R^2$ (all) $\uparrow$ & $\beta$ (degraded) $\uparrow$ & $R^2$ (degraded) $\uparrow$ \\
\midrule
Cosine (Distance) & $+0.322$ & \textbf{0.27} & $+0.598$ & \textbf{0.45} \\
MDS-BIC (Density) & $+0.228$ & 0.23 & $+0.309$ & 0.21 \\
IForest (Isolation) & $+0.160$ & 0.21 & $+0.185$ & 0.15 \\
ViM (Boundary / Hybrid) & $+0.184$ & 0.19 & $+0.230$ & 0.15 \\
Linear probe (Supervised) & $+0.321$ & 0.13 & $+0.745$ & 0.38 \\
MLP probe (Supervised) & $\mathbf{+0.361}$ & 0.15 & $\mathbf{+0.795}$ & 0.40 \\
\bottomrule
\end{tabular}
\end{table}

When do distance- and density-based detectors gain purchase? We examine this retrospectively using the per-setting log MAE gap, $\log_{10}(\mathrm{MAE}_{\mathrm{OOD}} / \mathrm{MAE}_{\mathrm{ID}})$. The intuition is that when the GNN's MAE degrades on OOD, the embedding has moved away from the training distribution in some measurable way, and detectors should be able to pick up on this displacement.

\looseness=-1
Figure~\ref{fig:auroc_vs_mae_gap_compacted} and Table~\ref{tab:regression_compacted} report OLS slopes $\beta$ and $R^2$ for one representative detector per family plus the supervised probes; the full per-detector grid is in Appendix Table~\ref{tab:regression_auroc_mae_gap}.
We find that distance and density detectors respond most strongly to this displacement, while isolation and boundary methods respond more weakly. In Figure~\ref{fig:auroc_vs_mae_gap_compacted}, the Cosine (Distance) and MDS-BIC (Density) panels show red regression lines from the degraded subset steeply sloping upward, meaning that AUROC rises with the MAE gap. The IForest (Isolation) and ViM (Boundary/Hybrid) panels show shallower red lines, indicating a weaker response to the regression failure signal. Cosine responds most strongly, with the steepest red line of any detector in the figure. Other detectors within each family follow these patterns (Appendix Table~\ref{tab:regression_auroc_mae_gap}).

However, this relationship is strongest when the GNN actually degrades on OOD samples. The blue regression lines, fit on all $N_{all} = 264$ settings, are visibly flatter than the red lines (Figure~\ref{fig:auroc_vs_mae_gap_compacted}), because settings with $\mathrm{MAE}_{\mathrm{OOD}} \leq \mathrm{MAE}_{\mathrm{ID}}$ add points with little shift for the MAE gap to track.

\subsection{Discussion: what MOOD reveals}
\label{sec:eval_discussion}

\looseness=-1
The current post-hoc OOD detection toolkit does not transfer to
materials GNNs, and its aggregate near chance performance hides two
different failure modes. In method-bottlenecked settings, frozen
embeddings contain ID/OOD signal, but unsupervised detectors fail to
extract it. In encoder-bottlenecked settings, even supervised probes
cannot recover ID/OOD signal, indicating that the relevant shift is not
encoded in the representation. A linear probe tests whether separable
OOD signal exists in the embedding, while the OOD-to-ID MAE ratio tests
whether the GNN's prediction error worsens under the shift. Together,
these diagnostics show whether failure comes from the detector or the encoder.

\looseness=-1
These two failure modes are physically interpretable. In the method-bottlenecked
tasks (e.g., dielectric/refractive index), the OOD
splits often change features that are also useful for property
prediction; refractive index depends strongly on composition-driven
electronic structure
\citep{shannon1993dielectric,petousis2017dielectric}. Thus, when MOOD applies compositional or
structural splits to these tasks, the GNN has reason to preserve the
same chemical or structural information that separates ID from OOD
samples. Supervised probes can recover this signal, even when current
unsupervised detectors cannot. By contrast, in the encoder-bottlenecked tasks (e.g., log\_kvrh/bulk modulus), accurate prediction can rely on more aggregated physical
signals such as mean bond stiffness and density for bulk modulus
\citep{deJong2015elastic}. These coarse signals may suffice for accurate property prediction but
may not preserve the composition or symmetry details needed to
separate ID from OOD samples.

\looseness=-1 A composition-only analysis supports this reading. On the encoder-bottlenecked tasks, probe AUROC stays within $0.02$ of chance even when ID and OOD materials differ substantially in Magpie compositional features~\citep{ward2016magpie}, so the shift exists in the input but does not survive the GNN's property prediction objective (Appendix~\ref{app:chemistry}).

\looseness=-1 The two regimes imply different next steps. Method-bottlenecked settings
call for better post-hoc detectors; the signal is present, but current
unsupervised methods do not extract it reliably. The method-bottleneck
subset (dielectric and log\_gvrh; 116 settings) is therefore the most
appropriate testbed for future detector work. Encoder-bottlenecked
settings instead call for better representation learning; if the frozen
GNN embedding does not encode the shift, a post-hoc detector has little
signal to use. Promising directions include auxiliary objectives,
contrastive pretraining over composition or symmetry variation, and
equivariance-aware architectures that preserve OOD relevant structure.

\FloatBarrier
\section{Conclusion}
\label{sec:conclusion}

\looseness=-1 We introduce MOOD, a benchmark for evaluating post-hoc OOD detection
on GNNs trained for materials property prediction. MOOD spans 5
MatBench regression tasks, 4 GNN architectures, and 15 physically
motivated compositional and structural shifts, yielding $N_{all} = 264$
evaluation settings across 17 detection methods and up to 4{,}488 AUROC
measurements per seed ($13{,}345$ across the three seeds). Two supervised diagnostic probes bound the OOD signal decodable from
each frozen embedding, separating method from encoder bottlenecks.

\looseness=-1
Three findings emerge. First, the vision OOD post-hoc toolkit does not
transfer, as every unsupervised detector achieves a median AUROC within
$0.01$ of chance. Second, these failures fall into two groups. On method-bottlenecked
tasks (dielectric, log\_gvrh), probes find OOD signal that detectors
miss. On encoder-bottlenecked tasks (log\_kvrh, perovskites, phonons),
neither probes nor detectors beat chance. Third, on the degraded
subset where OOD error exceeds ID error, AUROC tracks the MAE gap
with Cosine reaching $\beta = 0.60$, $R^2 = 0.45$. 

\paragraph{Limitations and future work.} \looseness=-1
MOOD focuses on post-hoc detectors applied to frozen GNN embeddings. We
do not evaluate training-time methods (outlier exposure, energy-based
fine-tuning, contrastive pretraining), detectors based on intermediate
activations, uncertainty, or ensembles, or domain adaptation methods
with access to target domain data. Results may also differ for equivariant architectures such as MACE or
for classification objectives. Our compositional and structural splits are physically motivated but not exhaustive. Alternatives such as
SOAP-LOCO~\citep{tan2025matuq} or property driven splits may reveal
additional failure modes. MOOD also scores only GNN embeddings; running
detectors directly on composition or structure fingerprints such as
Magpie~\citep{ward2016magpie} or SOAP~\citep{bartok2013soap} is a
natural next step. 

\paragraph{Broader impact.} \looseness=-1 MOOD helps practitioners judge when post-hoc OOD detection can work, so less simulation and synthesis effort is wasted on predictions that silently fail. A method-bottleneck label means the signal exists, not that a detector is ready to deploy; no detector we tested reaches practical accuracy. Our split rules reflect the makeup of the Materials Project data, and the released embeddings carry any blind spots of the encoders that produced them.

\begin{ack}
Experiments were run on the Princeton Computer Science Department's Ionic cluster. Adji Bousso Dieng acknowledges support from the National Science Foundation, Office of Advanced Cyberinfrastructure (OAC) \#2118201. Amey P. Pasarkar is supported by an NSF-GRFP fellowship. The authors declare no competing interests.
\end{ack}

\paragraph{AI use statement.} \looseness=-1 Generative AI tools were used to assist with implementing detection methods, building the evaluation pipeline, preparing the released code repository, and editing the manuscript for clarity. They were not used to formulate hypotheses, design the methodology, analyze data, or interpret results. The authors reviewed and tested all AI assisted code, verified that it reproduces published performance metrics where available, and take full responsibility for the content of this work.

\bibliographystyle{plainnat}
\bibliography{main}

\newpage
\appendix
\providecommand{\appendixroot}{appendix/}
%
%
%
%

\providecommand{\appendixroot}{}

\appendix

\renewcommand{\thesection}{\Alph{section}}

\section{Dataset details}
\label{app:dataset_details}

\paragraph{Source data and licensing.}
All datasets come from MatBench~\citep{dunn2020matbench}, which
collects regression tasks from the Materials Project. We do not
redistribute the source structures; we load them through MatBench's
matminer loader. MatBench is released under the BSD license; the
underlying Materials Project structures are released under CC-BY-4.0.

\paragraph{Selected tasks.}
We pick 5 of the 10 MatBench regression tasks: matbench\_dielectric
(refractive index), matbench\_log\_gvrh (log shear modulus),
matbench\_log\_kvrh (log bulk modulus), matbench\_perovskites
(formation energy), and matbench\_phonons (highest optical phonon frequency).
We chose these because they take crystal structures as input and are
big enough to support OOD splits with at least 50 minority class
samples. See Table~\ref{tab:dataset_stats}.

\begin{table}[t]
\centering
\small
\setlength{\tabcolsep}{4pt}
\caption{\textbf{Per-dataset benchmark statistics.} For each MatBench task we report the prediction target, total task size, number of OOD splits retained after the $\geq 50$ minority class filter, number of architectures evaluated, total number of settings (arch $\times$ split), and mean balanced ID/OOD test sizes across settings.}
\label{tab:dataset_stats}
\begin{tabular}{lllrrrrrr}
\toprule
Dataset & Target & MatBench $n$ & Splits & Archs & Cells & Mean $|\mathrm{ID}_\mathrm{test}|$ & Mean $|\mathrm{OOD}_\mathrm{test}|$ \\
\midrule
dielectric & refractive index & 4,764 & 14 & 4 & 56 & 361 & 361 \\
log\_gvrh & log shear modulus & 10,987 & 15 & 4 & 60 & 691 & 691 \\
log\_kvrh & log bulk modulus & 10,987 & 15 & 4 & 60 & 712 & 712 \\
perovskites & formation energy & 18,928 & 9 & 4 & 36 & 1,274 & 1,274 \\
phonons & phonon frequency & 1,265 & 13 & 4 & 52 & 98 & 98 \\
\bottomrule
\end{tabular}
\end{table}

\paragraph{Positioning relative to prior MatBench work.}
MatBench leaderboards report 5 fold cross-validation MAE on ID data
only; no public MatBench benchmark measures OOD MAE under
deterministic compositional or structural splits at the scale we
report here. We release our 264 setting ID/OOD MAE matrix
(Tables~\ref{tab:id_mae_per_arch} and~\ref{tab:per_setting_mae}) as a
separate artifact for future work on training time OOD interventions.

\paragraph{OOD split construction.}
For each dataset we build 15 binary splits: 6 compositional splits
(carbide / chalcogenide / halide / intermetallic / nitride / oxide)
and 9 structural splits (cubic / hexagonal / tetragonal /
orthorhombic / monoclinic / triclinic crystal systems /
high-spacegroup / low-spacegroup / perovskite rules). Each split
rule is a fixed function of a structure and is documented in the
data/ood\_splits.py file. The 90/10 ID train/test partition uses
random seed 42; the OOD test set is balanced to the same size as the
ID test set by stratified subsampling (also random seed 42). All
splits can be reproduced from a structure object and a seed.

\paragraph{Split rule definitions.}
Each of the 15 split rules is a fixed function of a pymatgen Structure
object. Table~\ref{tab:split_rule_definitions} lists them with
plain-language definitions; full implementations are in
data/ood\_splits.py. A spacegroup number is an integer from 1 to 230
(from the International Tables for Crystallography) that classifies a
crystal's symmetry, where higher means more symmetric. ABX$_3$ is a
chemical formula template with two cations A and B and an anion X in
1:1:3 ratio, the standard perovskite family.

\begin{table}[h]
  \centering
  \small
  \caption{\textbf{OOD split rule definitions.} Each rule maps a structure to a binary OOD label.}
  \label{tab:split_rule_definitions}
  \begin{tabular}{@{}l l p{6.5cm}@{}}
    \toprule
    Split rule & Type & Definition \\
    \midrule
    comp\_oxide        & composition & structure contains at least one O atom \\
    comp\_halide       & composition & structure contains at least one F/Cl/Br/I atom \\
    comp\_chalcogenide & composition & structure contains at least one S/Se/Te atom \\
    comp\_nitride      & composition & structure contains at least one N atom \\
    comp\_carbide      & composition & structure contains at least one C atom \\
    comp\_intermetallic& composition & structure contains only metallic elements \\
    \midrule
    struct\_cubic        & structure & spacegroup $\in [195, 230]$ \\
    struct\_hexagonal    & structure & spacegroup $\in [168, 194]$ \\
    struct\_tetragonal   & structure & spacegroup $\in [75, 142]$ \\
    struct\_orthorhombic & structure & spacegroup $\in [16, 74]$ \\
    struct\_monoclinic   & structure & spacegroup $\in [3, 15]$ \\
    struct\_triclinic    & structure & spacegroup $\in [1, 2]$ \\
    struct\_high\_spacegroup & structure & spacegroup $\geq 195$ \\
    struct\_low\_spacegroup  & structure & spacegroup $\leq 15$ \\
    struct\_perovskite   & structure & composition matches ABX$_3$ template \\
    \bottomrule
  \end{tabular}
\end{table}

\paragraph{Edge cases and excluded settings.}
We apply one filter across all analyses (Figures 2--5, Tables 1--4),
which produces the $264 / 300$ retained-settings count
(Table~\ref{tab:dropped_splits}), and handle one training failure:
\begin{itemize}[noitemsep,topsep=0pt]
    \item \textbf{Minority class size $<$ 50.} Splits with fewer than 50
        minority class samples are dropped at construction (9 splits;
        full list in Table~\ref{tab:dropped_splits}).
    \item \textbf{Encoder collapse.} Seven CGCNN/phonons encoders collapse to constant embeddings under seed 42
        (variance below $10^{-10}$; full list in
        Table~\ref{tab:dropped_splits}). All seven
        train normally under seeds 123 and 456, so the collapse is a property of the seed, not of
        CGCNN. We retain these settings and average over the two valid seeds, bringing the
        benchmark from 257 to 264 settings.
\end{itemize}

\begin{table}[t]
\centering
\small
\setlength{\tabcolsep}{8pt}
\caption{\textbf{Excluded splits and seed 42 collapses.} One filter reduces the candidate $5 \times 15 \times 4 = 300$ settings to the $264$ reported in the main paper. Top: splits dropped at construction because the minority OOD class has fewer than 50 samples in the source dataset ($N=9$ splits, all 4 architectures). Bottom: CGCNN/phonons encoders that collapse to constant embeddings under seed 42 only ($N=7$ settings); these train normally under seeds 123 and 456, are retained, and are averaged over the two valid seeds.}
\label{tab:dropped_splits}
\begin{tabular}{lllll}
\toprule
Dataset & Split & Arch & Reason & Details \\
\midrule
\multicolumn{5}{l}{\textit{Splits dropped at construction (minority class $<$ 50):}} \\
dielectric & comp\_carbide & all & minority class $<$ 50 & $N<50$ \\
perovskites & comp\_carbide & all & minority class $<$ 50 & $N<50$ \\
perovskites & comp\_intermetallic & all & minority class $<$ 50 & $N<50$ \\
perovskites & struct\_hexagonal & all & minority class $<$ 50 & $N<50$ \\
perovskites & struct\_low\_spacegroup & all & minority class $<$ 50 & $N<50$ \\
perovskites & struct\_monoclinic & all & minority class $<$ 50 & $N<50$ \\
perovskites & struct\_triclinic & all & minority class $<$ 50 & $N<50$ \\
phonons & comp\_carbide & all & minority class $<$ 50 & $N<50$ \\
phonons & struct\_triclinic & all & minority class $<$ 50 & $N<50$ \\
\addlinespace[2pt]
\multicolumn{5}{l}{\textit{Cells with seed 42 encoder collapse (retained, averaged over seeds 123, 456):}} \\
phonons & comp\_chalcogenide & cgcnn & seed 42 collapse & $\mathrm{var}(z) < 10^{-10}$ \\
phonons & comp\_halide & cgcnn & seed 42 collapse & $\mathrm{var}(z) < 10^{-10}$ \\
phonons & comp\_nitride & cgcnn & seed 42 collapse & $\mathrm{var}(z) < 10^{-10}$ \\
phonons & struct\_cubic & cgcnn & seed 42 collapse & $\mathrm{var}(z) < 10^{-10}$ \\
phonons & struct\_high\_spacegroup & cgcnn & seed 42 collapse & $\mathrm{var}(z) < 10^{-10}$ \\
phonons & struct\_low\_spacegroup & cgcnn & seed 42 collapse & $\mathrm{var}(z) < 10^{-10}$ \\
phonons & struct\_perovskite & cgcnn & seed 42 collapse & $\mathrm{var}(z) < 10^{-10}$ \\
\bottomrule
\end{tabular}
\end{table}

\begin{figure}[!htbp]
  \centering
  \includegraphics[width=\linewidth]{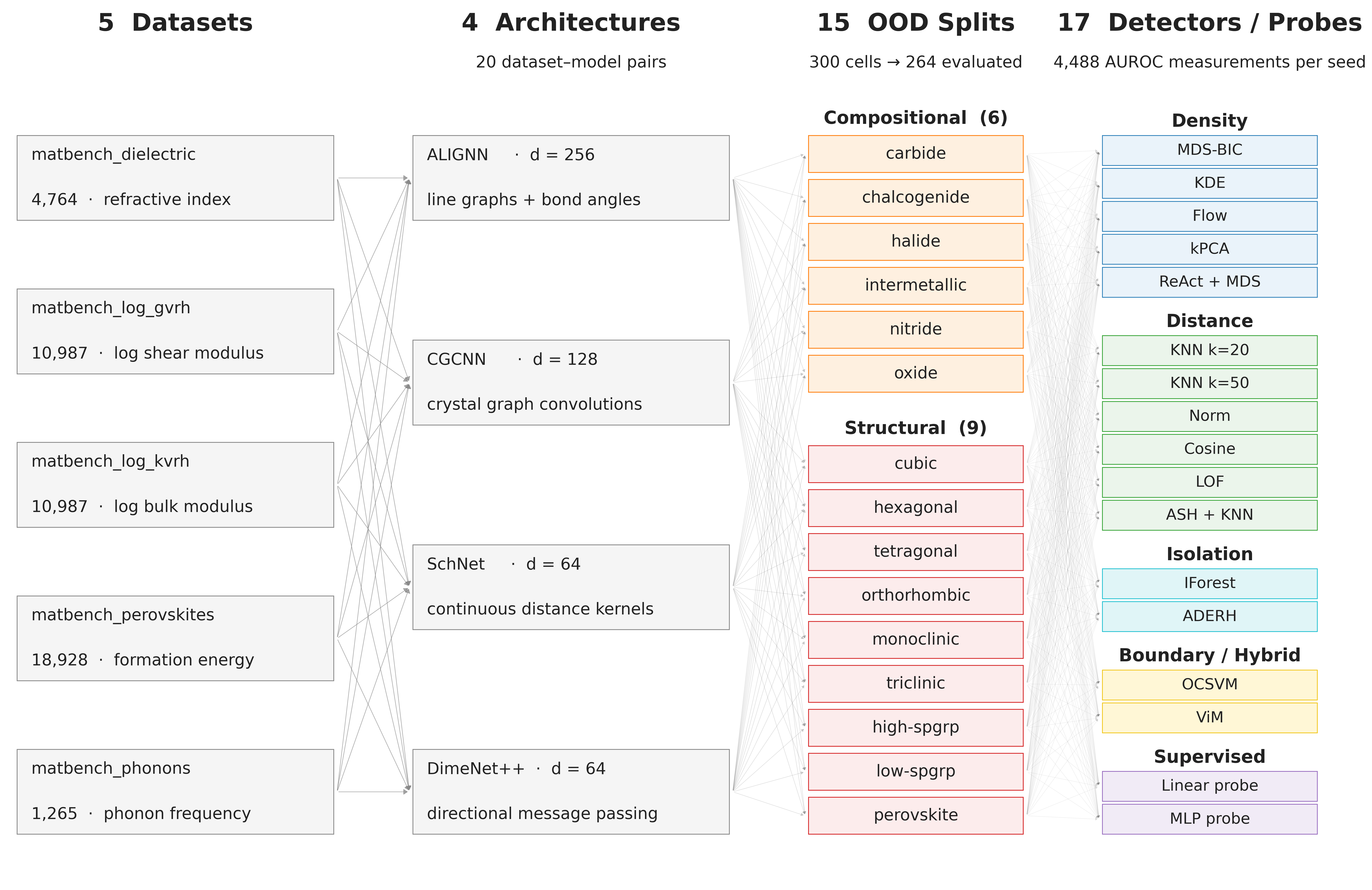}
  \caption{\textbf{MOOD benchmark axes.} 5 MatBench datasets, 4 GNN
  architectures, 15 OOD splits (6 compositional, 9 structural), and 17
  detectors across 5 families. The full grid yields 300 candidate
  (dataset, architecture, split) settings; 264 are kept after dropping
  splits with fewer than 50 minority class samples. Seven
  CGCNN/phonons settings collapse under seed 42 only and are averaged over
  the two valid seeds. Running 17
  detection methods on each setting yields $4{,}488$ AUROC measurements per seed.}
  \label{fig:benchmark_overview}
\end{figure}

\paragraph{Benchmark overview schematic.}
Figure~\ref{fig:benchmark_overview} shows the four benchmark axes
(datasets, architectures, OOD splits, detectors) and the
setting retention filtering described above.

\FloatBarrier

\section{Algorithm and baseline details}
\label{app:algorithms}
\label{app:detector_details}

We give implementation details for the 17 detection methods evaluated
in MOOD, one per item below. All detectors run on penultimate layer
GNN embeddings extracted from the trained encoder; the encoder is
held fixed during detection. Notation: $z \in \mathbb{R}^d$ is a test
embedding, $\mathrm{train}$ is the set of ID training embeddings, and
``higher score = more ID-like'' is the convention throughout (we flip
signs where the natural quantity goes the other way).

\paragraph{Distance-based detectors.}
\begin{itemize}[noitemsep,topsep=0pt]
    \item \textbf{KNN ($k = 20$ / $k = 50$).} Mean Euclidean distance to
        the $k$ nearest training neighbours, computed on $L_2$-normalised
        embeddings; sign-flipped so smaller distance scores higher.
    \item \textbf{Embedding norm.} $-\|z\|_2$, so smaller norms score as
        more ID-like. This sign is a fixed convention, opposite to that of
        \citet{yu2023featurenorm}; the detector performs near chance in our
        regression GNNs (mean AUROC $0.500$), so the choice of sign does not
        affect our conclusions.
    \item \textbf{Cosine.} Maximum cosine similarity to any training
        embedding,
        $\max_{z' \in \mathrm{train}} \langle z / \|z\|, z' / \|z'\| \rangle$.
        Computed once between the test point and every training
        embedding (not just the nearest in Euclidean space); higher = more
        ID-like.
    \item \textbf{LOF}~\citep{breunig2000lof}. Local Outlier Factor with
        $k = 20$ neighbours and the standard local-reachability-density
        formulation.
    \item \textbf{ASH$+$KNN}~\citep{djurisic2023ash}. Drop the lowest
        $90\%$ of feature magnitudes (90th-percentile threshold), then
        apply $k = 20$ KNN on the remaining embeddings.
\end{itemize}

\paragraph{Density-based detectors.}
\begin{itemize}[noitemsep,topsep=0pt]
    \item \textbf{MDS-BIC.} Mahalanobis distance under a $K$-component
        Gaussian mixture; $K \in \{1, 2, 4, 8, 16, 32, 64, 128\}$ chosen by the
        Bayesian information criterion on the ID training set.
    \item \textbf{Gaussian KDE.} Kernel density estimate with Scott's rule
        for bandwidth; OOD score is negative log-density.
    \item \textbf{Real-NVP}~\citep{dinh2017realnvp}. Normalising flow with
        $8$ coupling layers, hidden width $128$, trained $200$ epochs at
        learning rate $10^{-3}$, batch $256$, NLL loss. OOD score is
        the test point negative log-likelihood.
    \item \textbf{Kernel PCA reconstruction error}~\citep{hoffmann2007kpca}.
        RBF kernel with $\gamma = 2$, $50$ components; OOD score is the
        residual norm after reconstruction in feature space.
    \item \textbf{ReAct$+$MDS}~\citep{sun2021react}. Cap activations
        at the 90th percentile, then apply single-component Mahalanobis on
        the capped embeddings.
\end{itemize}

\paragraph{Isolation detectors.}
\begin{itemize}[noitemsep,topsep=0pt]
    \item \textbf{Isolation Forest}~\citep{liu2008iforest}. $100$
        estimators with the default subsample size
        $\min(256, n_{\mathrm{train}})$; OOD score is the (negative)
        average path length.
    \item \textbf{ADERH}~\citep{yang2024aderh}. Anomaly Detection by an
        Ensemble of Random pairs of Hyperspheres. Implemented per the
        published reference; no hyperparameters tuned on our data.
\end{itemize}

\paragraph{Boundary/hybrid detectors.}
\begin{itemize}[noitemsep,topsep=0pt]
    \item \textbf{One-class SVM}~\citep{scholkopf1999ocsvm}. RBF kernel
        with $\gamma =$ \texttt{'scale'} (sklearn default, $1/(d\cdot\mathrm{Var}(X))$)
        and $\nu = 0.1$; OOD score is the (negative)
        decision function.
    \item \textbf{ViM}~\citep{wang2022vim}. Used as a classifier-free
        residual-norm score: fit PCA on the centred ID training
        embeddings, keep the top components covering $95\%$ of the
        training variance, score a test embedding by the (negative) norm
        of its residual after reconstruction in that subspace. The
        original ViM combines this feature-space residual with a
        logit-space ``virtual logit'' term; we drop the logit term
        because our regression GNNs produce a single scalar output
        rather than a class manifold. See
        Appendix~\ref{app:detector_insights} for more.
\end{itemize}

\paragraph{Supervised diagnostic probes.}
\begin{itemize}[noitemsep,topsep=0pt]
    \item \textbf{Linear probe.} $L_2$-regularised logistic regression
        on the ID/OOD label, $C = 1.0$, LBFGS, max $1000$ iterations,
        inputs standardised per fold.
    \item \textbf{MLP probe.} Two hidden layers (widths $256$ and $128$,
        ReLU); Adam at learning rate $10^{-3}$, batch $256$, max $500$
        iterations; $L_2$ weight decay $\alpha = 1.0$ matched to the
        linear probe's $C = 1.0$; early stopping on a $10\%$ validation
        fold with patience $20$.
\end{itemize}
We evaluate both probes under 5 fold stratified cross-validation, with
mean AUROC across folds reported per setting. They are diagnostic
ceilings, not deployable detectors: they need OOD labels at fit time
and bound the OOD signal that a linear or MLP classifier can decode from the frozen embedding.

\paragraph{GNN backbones.} We use the reference implementations of
ALIGNN~\citep{choudhary2021alignn} ($d = 256$), CGCNN~\citep{xie2018cgcnn}
($d = 128$), SchNet~\citep{schutt2017schnet} ($d = 64$), and DimeNet++
\citep{gasteiger2020dimenet} ($d = 64$). We train each backbone on the
90/10 random partition of the ID set with default hyperparameters from
the original implementations.

\FloatBarrier

\section{Implementation details and reproducibility}
\label{app:reproducibility}

\paragraph{Code release.}
All training, embedding-extraction, detector-evaluation, and
figure-generation code is publicly released at
\url{https://github.com/vertaix/mood-benchmark}. The README lists the
command that regenerates each main-paper table and figure (python reproduce.py).

\paragraph{Pre-extracted embeddings.}
Penultimate-layer GNN embeddings for every retained (architecture,
dataset, split) setting are released on HuggingFace
(\url{https://huggingface.co/datasets/shirlyu/mood-benchmark}).
Researchers can reproduce all detector results without re-training the
GNNs by downloading this archive (about 3.7 GB).

\paragraph{Software versions.}
Pinned versions are tracked in requirements.txt in the code release.

\paragraph{Random seeds.}
Every random step uses an explicit seed:
\begin{itemize}[noitemsep,topsep=0pt]
    \item Split construction (scripts/generate\_physics\_splits.py): seed 42 for
        the 90/10 ID train/test partition and the balanced subsampling.
    \item Linear and MLP probe CV (scripts/run\_linear\_probe.py,
        scripts/run\_mlp\_probe.py): seed 42 for both the balanced ID/OOD
        subsample and StratifiedKFold with shuffling.
    \item Permutation test (analysis/probe\_permutation\_test.py): random state
        42 for sklearn's permutation\_test\_score (the balanced subsample
        and 5-fold CV). The test is run separately for each encoder seed---two
        seeds for the seven collapsed CGCNN/phonons settings---and per-dataset
        rejection rates and $z$-scores are reported as mean $\pm$ std across seeds.
    \item GNN training: three seeds (42, 123, 456) per
        (architecture, dataset, split) setting; reported numbers are mean
        $\pm$ std across seeds.
\end{itemize}

\paragraph{Compute used.} Training all $264 \times 3 - 7 = 785$ encoders (264 settings over three
seeds, minus the seven CGCNN/phonons settings that collapse under seed 42) took about 288 GPU hours on a
single node NVIDIA A100 cluster. Detector evaluation is CPU only: the
full 15 detector $\times$ 264 setting sweep, run across all three encoder seeds
(including the per-seed linear probe permutation test at $B = 100$ shuffles),
took about 740 CPU hours. Re-using the released embeddings, evaluating one new
detector on the full benchmark takes under one CPU hour.

\FloatBarrier

\section{Complete numerical results}
\label{app:numerical_results}

This section contains the full per-architecture, per-split-type, and
per-detector breakdowns underlying Figures~2--5 and Tables~1--2 of the
main paper. Each per-table preamble points back to the main-paper
counterpart where applicable.

\paragraph{Per-detector performance summary (full version).}
Table~\ref{tab:per_detector_summary_full} adds two columns to the main
paper Table~\ref{tab:per_detector_summary}: mean AUROC
(degraded), restricted to the $N_{deg} = 153$ settings where
seed mean OOD MAE $>$ seed mean ID MAE (the regime where the GNN degrades on OOD),
and probe gap, the mean of (linear probe AUROC $-$ this
detector's AUROC) across all settings. Two patterns emerge that
aren't visible in the compact main paper table. First,
every detector improves on the degraded subset, by $+0.002$ to $+0.018$ (e.g., Cosine:
$0.531 \to 0.549$, IForest: $0.502 \to 0.509$): restricting to the
regime where embeddings register the shift gives every method a small
but consistent boost. Second, the probe gap is non-trivial even for the
best detector: Cosine recovers only about three quarters of the linear
probe's above chance AUROC---$72\%$ overall and $75\%$ on the degraded subset.

\begin{table}[t]
\centering
\small
\setlength{\tabcolsep}{3pt}
\caption{\textbf{Per-detector performance summary} (full version with degraded subset and probe gap columns; the main paper Table~\ref{tab:per_detector_summary} omits these for width; central values are means over three encoder seeds). Mean / Median AUROC: across all $N_{all}=264$ settings. Mean AUROC (degraded): restricted to the $N_{deg}=153$ settings where seed mean OOD MAE $>$ seed mean ID MAE. Probe gap: mean per-setting (linear probe AUROC $-$ this AUROC); smaller is better. Win count: ties split equally. \textbf{Bold} is the best value per column among unsupervised detectors. $\uparrow$ indicates higher is better; $\downarrow$ indicates lower is better.}
\label{tab:per_detector_summary_full}
\begin{tabular}{llcccccc}
\toprule
Family & Detector & Mean $\uparrow$ & Median $\uparrow$ & Mean (deg.) $\uparrow$ & Probe gap $\downarrow$ & Wins $\uparrow$ & \%$>$0.55 $\uparrow$ \\
\midrule
\multirow{6}{*}{Distance} & KNN-k20-norm & 0.512 & 0.504 & 0.523 & +0.031 & 13.2 & 10.4\% \\
 & KNN-k50-norm & 0.508 & 0.503 & 0.518 & +0.034 & 7.7 & 9.3\% \\
 & Norm & 0.500 & 0.500 & 0.504 & +0.043 & 23.3 & 6.1\% \\
 & Cosine & \textbf{0.531} & \textbf{0.507} & \textbf{0.549} & \textbf{+0.012} & \textbf{54.8} & \textbf{13.5\%} \\
 & LOF & 0.526 & 0.505 & 0.542 & +0.017 & 30.2 & 13.1\% \\
 & ASH+KNN & 0.516 & 0.505 & 0.530 & +0.026 & 30.8 & 11.3\% \\
\cmidrule(lr){1-8}
\multirow{5}{*}{Density} & MDS-BIC & 0.513 & 0.504 & 0.526 & +0.029 & 3.7 & 8.5\% \\
 & KDE & 0.518 & 0.504 & 0.531 & +0.025 & 16.7 & 10.2\% \\
 & Flow & 0.520 & 0.503 & 0.535 & +0.023 & 17.3 & 9.0\% \\
 & kPCA & 0.512 & 0.503 & 0.524 & +0.030 & 5.7 & 9.0\% \\
 & ReAct+MDS & 0.512 & 0.503 & 0.525 & +0.030 & 6.0 & 8.0\% \\
\cmidrule(lr){1-8}
\multirow{2}{*}{Isolation} & IForest & 0.502 & 0.500 & 0.509 & +0.040 & 11.7 & 7.4\% \\
 & ADERH & 0.497 & 0.500 & 0.501 & +0.045 & 11.2 & 4.7\% \\
\cmidrule(lr){1-8}
\multirow{2}{*}{\shortstack{Boundary/\\Hybrid}} & OCSVM & 0.496 & 0.498 & 0.498 & +0.047 & 19.3 & 7.4\% \\
 & ViM & 0.509 & 0.503 & 0.519 & +0.034 & 10.2 & 8.5\% \\
\midrule
\multirow{2}{*}{Supervised} & Linear probe & 0.543 & 0.500 & 0.565 & -- & -- & 16.4\% \\
 & MLP probe & 0.542 & 0.496 & 0.567 & -- & -- & 15.5\% \\

\bottomrule
\end{tabular}
\end{table}

\begin{figure}[!htbp]
  \centering
  \includegraphics[width=\linewidth]{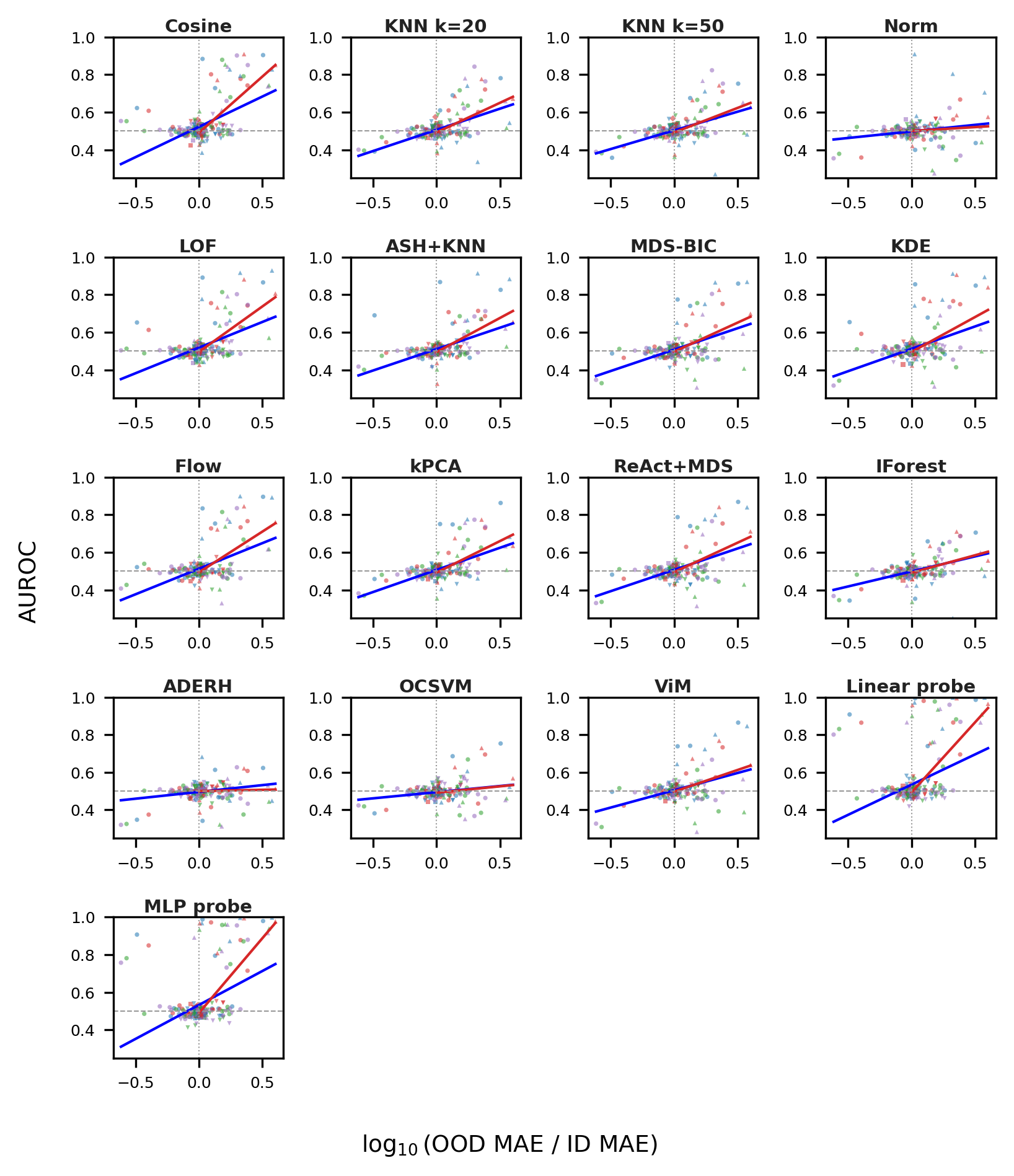}
  \caption{\textbf{Detector AUROC vs $\log_{10}(\text{OOD MAE} / \text{ID MAE})$,
  one panel per detector.} Full $4\times5$ grid ($17$ panels) corresponding to the
  compacted main-paper Figure~\ref{fig:auroc_vs_mae_gap_compacted} (one
  panel per family). Blue regression lines fit all $N_{all}= 264$ settings; red
  regression lines fit the degraded subset ($N_{deg} = 153$, seed mean OOD MAE
  $>$ seed mean ID MAE).}
  \label{fig:auroc_vs_mae_gap_full}
\end{figure}

\paragraph{Held-out ID vs OOD MAE per architecture.}
Table~\ref{tab:id_mae_per_arch} reports the held out (test fold) regression
error of each architecture on each dataset, averaged across the OOD splits
retained for that dataset. The OOD/ID gap is the independent variable in
Figure~\ref{fig:auroc_vs_mae_gap_full} and
Table~\ref{tab:regression_auroc_mae_gap}.

\begin{table}[t]
\centering
\small
\setlength{\tabcolsep}{3pt}
\caption{\textbf{Mean ID and OOD MAE per dataset and architecture (mean $\pm$ std over three encoder seeds).} Averaged across the OOD splits retained for each dataset. $\downarrow$ indicates lower is better.}
\label{tab:id_mae_per_arch}
\resizebox{\linewidth}{!}{%
\begin{tabular}{lrrrrrrrr}
\toprule
 & \multicolumn{2}{c}{Alignn} & \multicolumn{2}{c}{CGCNN} & \multicolumn{2}{c}{SchNet} & \multicolumn{2}{c}{DimeNet} \\
\cmidrule(lr){2-3}\cmidrule(lr){4-5}\cmidrule(lr){6-7}\cmidrule(lr){8-9}
Dataset & ID $\downarrow$ & OOD $\downarrow$ & ID $\downarrow$ & OOD $\downarrow$ & ID $\downarrow$ & OOD $\downarrow$ & ID $\downarrow$ & OOD $\downarrow$ \\
\midrule
dielectric & 0.447 {\tiny$\pm$0.020} & 0.480 {\tiny$\pm$0.018} & 0.567 {\tiny$\pm$0.007} & 0.605 {\tiny$\pm$0.021} & 0.417 {\tiny$\pm$0.028} & 0.463 {\tiny$\pm$0.010} & 0.430 {\tiny$\pm$0.025} & 0.476 {\tiny$\pm$0.034} \\
log\_gvrh & 0.074 {\tiny$\pm$0.002} & 0.098 {\tiny$\pm$0.001} & 0.097 {\tiny$\pm$0.002} & 0.128 {\tiny$\pm$0.007} & 0.094 {\tiny$\pm$0.003} & 0.118 {\tiny$\pm$0.004} & 0.098 {\tiny$\pm$0.006} & 0.124 {\tiny$\pm$0.008} \\
log\_kvrh & 0.060 {\tiny$\pm$0.003} & 0.061 {\tiny$\pm$0.003} & 0.079 {\tiny$\pm$0.005} & 0.079 {\tiny$\pm$0.006} & 0.070 {\tiny$\pm$0.002} & 0.071 {\tiny$\pm$0.002} & 0.069 {\tiny$\pm$0.001} & 0.069 {\tiny$\pm$0.000} \\
perovskites & 0.034 {\tiny$\pm$0.001} & 0.033 {\tiny$\pm$0.001} & 0.069 {\tiny$\pm$0.005} & 0.068 {\tiny$\pm$0.005} & 0.046 {\tiny$\pm$0.005} & 0.046 {\tiny$\pm$0.005} & 0.045 {\tiny$\pm$0.007} & 0.045 {\tiny$\pm$0.007} \\
phonons & 34.9 {\tiny$\pm$8.2} & 38.3 {\tiny$\pm$7.6} & 78.4 {\tiny$\pm$3.6} & 85.8 {\tiny$\pm$1.6} & 58.3 {\tiny$\pm$4.3} & 65.2 {\tiny$\pm$4.3} & 58.1 {\tiny$\pm$5.7} & 66.2 {\tiny$\pm$8.4} \\
\bottomrule
\end{tabular}}
\end{table}

\paragraph{Held-out ID and OOD MAE per setting.}
Table~\ref{tab:per_setting_mae} reports the same numbers at the level
of individual (task, split, architecture) settings---one row per retained
setting, $N_{all} = 264$---which is the data input that justifies the
sample size of the central regression in
Table~\ref{tab:regression_auroc_mae_gap}.

{\setlength{\LTcapwidth}{\linewidth}
\begin{longtable}{@{\extracolsep{\fill}}llrrr@{}}
\caption{\textbf{Per-setting held-out ID and OOD MAE (mean $\pm$ std over three encoder seeds).} Each row is one retained (dataset, split, arch) setting ($N=264$). $\log_{10}$ gap $=\log_{10}(\text{seed mean OOD MAE}/\text{seed mean ID MAE})$.}
\label{tab:per_setting_mae} \\
\toprule
Dataset & Split & Arch & ID MAE & OOD MAE \\
\midrule
\endfirsthead
\multicolumn{5}{l}{\small\itshape (continued)} \\
\toprule
Dataset & Split & Arch & ID MAE & OOD MAE \\
\midrule
\endhead
\midrule
\multicolumn{5}{r}{\small\itshape (continued)} \\
\endfoot
\bottomrule
\endlastfoot
dielectric & comp\_chalcogenide & alignn & 0.4153{\tiny$\pm$0.0313} & 0.3598{\tiny$\pm$0.0395} \\
 &  & cgcnn & 0.5958{\tiny$\pm$0.0585} & 0.5928{\tiny$\pm$0.1980} \\
 &  & dimenet & 0.6899{\tiny$\pm$0.3254} & 0.3388{\tiny$\pm$0.0031} \\
 &  & schnet & 0.3327{\tiny$\pm$0.0181} & 0.5772{\tiny$\pm$0.0804} \\
 & comp\_halide & alignn & 0.4861{\tiny$\pm$0.0517} & 1.5563{\tiny$\pm$0.0865} \\
 &  & cgcnn & 0.5829{\tiny$\pm$0.0062} & 1.2402{\tiny$\pm$0.3578} \\
 &  & dimenet & 0.4748{\tiny$\pm$0.0079} & 1.1516{\tiny$\pm$0.2391} \\
 &  & schnet & 0.4611{\tiny$\pm$0.0362} & 1.0344{\tiny$\pm$0.0649} \\
 & comp\_intermetallic & alignn & 0.3751{\tiny$\pm$0.0276} & 0.5342{\tiny$\pm$0.0399} \\
 &  & cgcnn & 0.5907{\tiny$\pm$0.0869} & 0.6340{\tiny$\pm$0.0433} \\
 &  & dimenet & 0.3356{\tiny$\pm$0.0217} & 0.5186{\tiny$\pm$0.0044} \\
 &  & schnet & 0.3567{\tiny$\pm$0.0250} & 0.5434{\tiny$\pm$0.0186} \\
 & comp\_nitride & alignn & 0.3292{\tiny$\pm$0.0645} & 0.2553{\tiny$\pm$0.0168} \\
 &  & cgcnn & 0.4791{\tiny$\pm$0.1020} & 0.3821{\tiny$\pm$0.0683} \\
 &  & dimenet & 0.3038{\tiny$\pm$0.0468} & 0.2412{\tiny$\pm$0.0103} \\
 &  & schnet & 0.2674{\tiny$\pm$0.0093} & 0.2011{\tiny$\pm$0.0058} \\
 & comp\_oxide & alignn & 0.5089{\tiny$\pm$0.0189} & 0.5428{\tiny$\pm$0.0922} \\
 &  & cgcnn & 0.6761{\tiny$\pm$0.0171} & 0.8417{\tiny$\pm$0.0507} \\
 &  & dimenet & 0.5582{\tiny$\pm$0.0382} & 1.1110{\tiny$\pm$0.2785} \\
 &  & schnet & 0.5787{\tiny$\pm$0.0801} & 0.8817{\tiny$\pm$0.1868} \\
 & struct\_cubic & alignn & 0.4902{\tiny$\pm$0.0227} & 0.6567{\tiny$\pm$0.0718} \\
 &  & cgcnn & 0.4952{\tiny$\pm$0.0051} & 1.1952{\tiny$\pm$0.0768} \\
 &  & dimenet & 0.3821{\tiny$\pm$0.0091} & 0.6307{\tiny$\pm$0.0011} \\
 &  & schnet & 0.3803{\tiny$\pm$0.0144} & 0.6721{\tiny$\pm$0.0752} \\
 & struct\_hexagonal & alignn & 0.5692{\tiny$\pm$0.0324} & 0.3972{\tiny$\pm$0.0117} \\
 &  & cgcnn & 0.6212{\tiny$\pm$0.0110} & 0.5079{\tiny$\pm$0.0028} \\
 &  & dimenet & 0.5416{\tiny$\pm$0.0447} & 0.4218{\tiny$\pm$0.0127} \\
 &  & schnet & 0.5212{\tiny$\pm$0.0024} & 0.4019{\tiny$\pm$0.0046} \\
 & struct\_high\_spacegroup & alignn & 0.3448{\tiny$\pm$0.0276} & 0.4371{\tiny$\pm$0.0420} \\
 &  & cgcnn & 0.4167{\tiny$\pm$0.0176} & 0.5368{\tiny$\pm$0.0066} \\
 &  & dimenet & 0.2537{\tiny$\pm$0.0106} & 0.5386{\tiny$\pm$0.1293} \\
 &  & schnet & 0.2618{\tiny$\pm$0.0182} & 0.4395{\tiny$\pm$0.0356} \\
 & struct\_low\_spacegroup & alignn & 0.5645{\tiny$\pm$0.0716} & 0.3929{\tiny$\pm$0.0226} \\
 &  & cgcnn & 0.6127{\tiny$\pm$0.0285} & 0.4517{\tiny$\pm$0.0376} \\
 &  & dimenet & 0.5813{\tiny$\pm$0.1046} & 0.3522{\tiny$\pm$0.0019} \\
 &  & schnet & 0.4505{\tiny$\pm$0.0236} & 0.3497{\tiny$\pm$0.0069} \\
 & struct\_monoclinic & alignn & 0.2582{\tiny$\pm$0.0303} & 0.4716{\tiny$\pm$0.0215} \\
 &  & cgcnn & 0.3561{\tiny$\pm$0.0473} & 0.5177{\tiny$\pm$0.0310} \\
 &  & dimenet & 0.2242{\tiny$\pm$0.0165} & 0.3931{\tiny$\pm$0.0087} \\
 &  & schnet & 0.2343{\tiny$\pm$0.0081} & 0.3943{\tiny$\pm$0.0103} \\
 & struct\_orthorhombic & alignn & 0.5271{\tiny$\pm$0.0383} & 0.3159{\tiny$\pm$0.0196} \\
 &  & cgcnn & 0.7976{\tiny$\pm$0.1458} & 0.4954{\tiny$\pm$0.0771} \\
 &  & dimenet & 0.4116{\tiny$\pm$0.0299} & 0.3013{\tiny$\pm$0.0219} \\
 &  & schnet & 0.8029{\tiny$\pm$0.3031} & 0.2972{\tiny$\pm$0.0165} \\
 & struct\_perovskite & alignn & 0.5039{\tiny$\pm$0.0279} & 0.1635{\tiny$\pm$0.0103} \\
 &  & cgcnn & 0.5811{\tiny$\pm$0.0069} & 0.2331{\tiny$\pm$0.0176} \\
 &  & dimenet & 0.4460{\tiny$\pm$0.0256} & 0.1082{\tiny$\pm$0.0179} \\
 &  & schnet & 0.3998{\tiny$\pm$0.0051} & 0.1073{\tiny$\pm$0.0043} \\
 & struct\_tetragonal & alignn & 0.5807{\tiny$\pm$0.0147} & 0.3693{\tiny$\pm$0.0302} \\
 &  & cgcnn & 0.6798{\tiny$\pm$0.0484} & 0.4773{\tiny$\pm$0.0647} \\
 &  & dimenet & 0.5032{\tiny$\pm$0.0116} & 0.2959{\tiny$\pm$0.0059} \\
 &  & schnet & 0.5068{\tiny$\pm$0.0106} & 0.3334{\tiny$\pm$0.0368} \\
 & struct\_triclinic & alignn & 0.3027{\tiny$\pm$0.0435} & 0.2702{\tiny$\pm$0.0349} \\
 &  & cgcnn & 0.4494{\tiny$\pm$0.0261} & 0.3676{\tiny$\pm$0.0476} \\
 &  & dimenet & 0.3203{\tiny$\pm$0.0121} & 0.2672{\tiny$\pm$0.0081} \\
 &  & schnet & 0.2885{\tiny$\pm$0.0106} & 0.2436{\tiny$\pm$0.0135} \\
\addlinespace[2pt]
log\_gvrh & comp\_carbide & alignn & 0.0733{\tiny$\pm$0.0020} & 0.0831{\tiny$\pm$0.0023} \\
 &  & cgcnn & 0.0974{\tiny$\pm$0.0053} & 0.1057{\tiny$\pm$0.0043} \\
 &  & dimenet & 0.0967{\tiny$\pm$0.0077} & 0.1086{\tiny$\pm$0.0117} \\
 &  & schnet & 0.0934{\tiny$\pm$0.0049} & 0.1070{\tiny$\pm$0.0018} \\
 & comp\_chalcogenide & alignn & 0.0773{\tiny$\pm$0.0025} & 0.0718{\tiny$\pm$0.0015} \\
 &  & cgcnn & 0.1008{\tiny$\pm$0.0022} & 0.0929{\tiny$\pm$0.0008} \\
 &  & dimenet & 0.1017{\tiny$\pm$0.0065} & 0.0967{\tiny$\pm$0.0062} \\
 &  & schnet & 0.0962{\tiny$\pm$0.0060} & 0.0921{\tiny$\pm$0.0013} \\
 & comp\_halide & alignn & 0.0707{\tiny$\pm$0.0005} & 0.2662{\tiny$\pm$0.0086} \\
 &  & cgcnn & 0.0955{\tiny$\pm$0.0059} & 0.3824{\tiny$\pm$0.0620} \\
 &  & dimenet & 0.0947{\tiny$\pm$0.0033} & 0.3309{\tiny$\pm$0.0316} \\
 &  & schnet & 0.0874{\tiny$\pm$0.0035} & 0.3114{\tiny$\pm$0.0161} \\
 & comp\_intermetallic & alignn & 0.0799{\tiny$\pm$0.0030} & 0.0853{\tiny$\pm$0.0022} \\
 &  & cgcnn & 0.1025{\tiny$\pm$0.0090} & 0.1086{\tiny$\pm$0.0040} \\
 &  & dimenet & 0.1043{\tiny$\pm$0.0031} & 0.1199{\tiny$\pm$0.0037} \\
 &  & schnet & 0.0996{\tiny$\pm$0.0035} & 0.1061{\tiny$\pm$0.0007} \\
 & comp\_nitride & alignn & 0.0756{\tiny$\pm$0.0010} & 0.0717{\tiny$\pm$0.0039} \\
 &  & cgcnn & 0.0976{\tiny$\pm$0.0012} & 0.0966{\tiny$\pm$0.0008} \\
 &  & dimenet & 0.0929{\tiny$\pm$0.0054} & 0.0963{\tiny$\pm$0.0040} \\
 &  & schnet & 0.0937{\tiny$\pm$0.0047} & 0.0920{\tiny$\pm$0.0035} \\
 & comp\_oxide & alignn & 0.0710{\tiny$\pm$0.0003} & 0.1498{\tiny$\pm$0.0116} \\
 &  & cgcnn & 0.0961{\tiny$\pm$0.0022} & 0.2176{\tiny$\pm$0.0395} \\
 &  & dimenet & 0.0992{\tiny$\pm$0.0099} & 0.1666{\tiny$\pm$0.0331} \\
 &  & schnet & 0.0922{\tiny$\pm$0.0005} & 0.1486{\tiny$\pm$0.0155} \\
 & struct\_cubic & alignn & 0.0678{\tiny$\pm$0.0008} & 0.1186{\tiny$\pm$0.0006} \\
 &  & cgcnn & 0.0949{\tiny$\pm$0.0021} & 0.1325{\tiny$\pm$0.0007} \\
 &  & dimenet & 0.0946{\tiny$\pm$0.0022} & 0.1430{\tiny$\pm$0.0056} \\
 &  & schnet & 0.0943{\tiny$\pm$0.0053} & 0.1375{\tiny$\pm$0.0087} \\
 & struct\_hexagonal & alignn & 0.0726{\tiny$\pm$0.0034} & 0.0734{\tiny$\pm$0.0023} \\
 &  & cgcnn & 0.0935{\tiny$\pm$0.0004} & 0.0971{\tiny$\pm$0.0013} \\
 &  & dimenet & 0.0957{\tiny$\pm$0.0018} & 0.1010{\tiny$\pm$0.0016} \\
 &  & schnet & 0.0918{\tiny$\pm$0.0018} & 0.0951{\tiny$\pm$0.0014} \\
 & struct\_high\_spacegroup & alignn & 0.0833{\tiny$\pm$0.0021} & 0.0811{\tiny$\pm$0.0033} \\
 &  & cgcnn & 0.1042{\tiny$\pm$0.0024} & 0.1024{\tiny$\pm$0.0018} \\
 &  & dimenet & 0.1063{\tiny$\pm$0.0112} & 0.1069{\tiny$\pm$0.0084} \\
 &  & schnet & 0.1010{\tiny$\pm$0.0022} & 0.0982{\tiny$\pm$0.0012} \\
 & struct\_low\_spacegroup & alignn & 0.0701{\tiny$\pm$0.0027} & 0.0767{\tiny$\pm$0.0026} \\
 &  & cgcnn & 0.0947{\tiny$\pm$0.0027} & 0.0972{\tiny$\pm$0.0019} \\
 &  & dimenet & 0.0900{\tiny$\pm$0.0040} & 0.0986{\tiny$\pm$0.0038} \\
 &  & schnet & 0.0924{\tiny$\pm$0.0056} & 0.0964{\tiny$\pm$0.0045} \\
 & struct\_monoclinic & alignn & 0.0733{\tiny$\pm$0.0016} & 0.0742{\tiny$\pm$0.0034} \\
 &  & cgcnn & 0.0977{\tiny$\pm$0.0019} & 0.0969{\tiny$\pm$0.0008} \\
 &  & dimenet & 0.0917{\tiny$\pm$0.0037} & 0.0964{\tiny$\pm$0.0038} \\
 &  & schnet & 0.0939{\tiny$\pm$0.0016} & 0.0946{\tiny$\pm$0.0028} \\
 & struct\_orthorhombic & alignn & 0.0721{\tiny$\pm$0.0002} & 0.0744{\tiny$\pm$0.0004} \\
 &  & cgcnn & 0.0973{\tiny$\pm$0.0005} & 0.0991{\tiny$\pm$0.0006} \\
 &  & dimenet & 0.0991{\tiny$\pm$0.0086} & 0.1048{\tiny$\pm$0.0069} \\
 &  & schnet & 0.0950{\tiny$\pm$0.0055} & 0.0981{\tiny$\pm$0.0045} \\
 & struct\_perovskite & alignn & 0.0760{\tiny$\pm$0.0023} & 0.0802{\tiny$\pm$0.0008} \\
 &  & cgcnn & 0.0962{\tiny$\pm$0.0006} & 0.0975{\tiny$\pm$0.0038} \\
 &  & dimenet & 0.1026{\tiny$\pm$0.0087} & 0.0941{\tiny$\pm$0.0114} \\
 &  & schnet & 0.0976{\tiny$\pm$0.0011} & 0.0984{\tiny$\pm$0.0026} \\
 & struct\_tetragonal & alignn & 0.0764{\tiny$\pm$0.0026} & 0.0770{\tiny$\pm$0.0030} \\
 &  & cgcnn & 0.0962{\tiny$\pm$0.0008} & 0.0980{\tiny$\pm$0.0018} \\
 &  & dimenet & 0.1023{\tiny$\pm$0.0077} & 0.0979{\tiny$\pm$0.0074} \\
 &  & schnet & 0.0959{\tiny$\pm$0.0036} & 0.0962{\tiny$\pm$0.0024} \\
 & struct\_triclinic & alignn & 0.0723{\tiny$\pm$0.0026} & 0.0790{\tiny$\pm$0.0052} \\
 &  & cgcnn & 0.0929{\tiny$\pm$0.0017} & 0.0936{\tiny$\pm$0.0045} \\
 &  & dimenet & 0.0939{\tiny$\pm$0.0044} & 0.0922{\tiny$\pm$0.0050} \\
 &  & schnet & 0.0917{\tiny$\pm$0.0029} & 0.0967{\tiny$\pm$0.0060} \\
\addlinespace[2pt]
log\_kvrh & comp\_carbide & alignn & 0.0578{\tiny$\pm$0.0029} & 0.0586{\tiny$\pm$0.0021} \\
 &  & cgcnn & 0.0772{\tiny$\pm$0.0063} & 0.0824{\tiny$\pm$0.0091} \\
 &  & dimenet & 0.0649{\tiny$\pm$0.0024} & 0.0637{\tiny$\pm$0.0026} \\
 &  & schnet & 0.0700{\tiny$\pm$0.0035} & 0.0692{\tiny$\pm$0.0026} \\
 & comp\_chalcogenide & alignn & 0.0558{\tiny$\pm$0.0025} & 0.0589{\tiny$\pm$0.0028} \\
 &  & cgcnn & 0.0754{\tiny$\pm$0.0069} & 0.0782{\tiny$\pm$0.0054} \\
 &  & dimenet & 0.0652{\tiny$\pm$0.0035} & 0.0669{\tiny$\pm$0.0037} \\
 &  & schnet & 0.0665{\tiny$\pm$0.0024} & 0.0694{\tiny$\pm$0.0022} \\
 & comp\_halide & alignn & 0.0604{\tiny$\pm$0.0036} & 0.0591{\tiny$\pm$0.0034} \\
 &  & cgcnn & 0.0833{\tiny$\pm$0.0056} & 0.0779{\tiny$\pm$0.0052} \\
 &  & dimenet & 0.0691{\tiny$\pm$0.0023} & 0.0650{\tiny$\pm$0.0015} \\
 &  & schnet & 0.0731{\tiny$\pm$0.0031} & 0.0668{\tiny$\pm$0.0023} \\
 & comp\_intermetallic & alignn & 0.0664{\tiny$\pm$0.0055} & 0.0678{\tiny$\pm$0.0046} \\
 &  & cgcnn & 0.0830{\tiny$\pm$0.0056} & 0.0871{\tiny$\pm$0.0068} \\
 &  & dimenet & 0.0731{\tiny$\pm$0.0016} & 0.0806{\tiny$\pm$0.0018} \\
 &  & schnet & 0.0759{\tiny$\pm$0.0030} & 0.0789{\tiny$\pm$0.0025} \\
 & comp\_nitride & alignn & 0.0621{\tiny$\pm$0.0017} & 0.0638{\tiny$\pm$0.0040} \\
 &  & cgcnn & 0.0809{\tiny$\pm$0.0058} & 0.0811{\tiny$\pm$0.0051} \\
 &  & dimenet & 0.0668{\tiny$\pm$0.0046} & 0.0684{\tiny$\pm$0.0052} \\
 &  & schnet & 0.0691{\tiny$\pm$0.0005} & 0.0742{\tiny$\pm$0.0022} \\
 & comp\_oxide & alignn & 0.0646{\tiny$\pm$0.0047} & 0.0628{\tiny$\pm$0.0044} \\
 &  & cgcnn & 0.0833{\tiny$\pm$0.0070} & 0.0833{\tiny$\pm$0.0060} \\
 &  & dimenet & 0.0722{\tiny$\pm$0.0014} & 0.0720{\tiny$\pm$0.0018} \\
 &  & schnet & 0.0718{\tiny$\pm$0.0028} & 0.0720{\tiny$\pm$0.0035} \\
 & struct\_cubic & alignn & 0.0634{\tiny$\pm$0.0039} & 0.0639{\tiny$\pm$0.0064} \\
 &  & cgcnn & 0.0796{\tiny$\pm$0.0082} & 0.0850{\tiny$\pm$0.0065} \\
 &  & dimenet & 0.0727{\tiny$\pm$0.0045} & 0.0745{\tiny$\pm$0.0024} \\
 &  & schnet & 0.0715{\tiny$\pm$0.0019} & 0.0758{\tiny$\pm$0.0018} \\
 & struct\_hexagonal & alignn & 0.0566{\tiny$\pm$0.0046} & 0.0572{\tiny$\pm$0.0030} \\
 &  & cgcnn & 0.0756{\tiny$\pm$0.0042} & 0.0761{\tiny$\pm$0.0057} \\
 &  & dimenet & 0.0730{\tiny$\pm$0.0050} & 0.0704{\tiny$\pm$0.0054} \\
 &  & schnet & 0.0689{\tiny$\pm$0.0020} & 0.0690{\tiny$\pm$0.0026} \\
 & struct\_high\_spacegroup & alignn & 0.0634{\tiny$\pm$0.0039} & 0.0639{\tiny$\pm$0.0064} \\
 &  & cgcnn & 0.0796{\tiny$\pm$0.0082} & 0.0850{\tiny$\pm$0.0065} \\
 &  & dimenet & 0.0736{\tiny$\pm$0.0019} & 0.0762{\tiny$\pm$0.0037} \\
 &  & schnet & 0.0700{\tiny$\pm$0.0020} & 0.0762{\tiny$\pm$0.0028} \\
 & struct\_low\_spacegroup & alignn & 0.0601{\tiny$\pm$0.0042} & 0.0621{\tiny$\pm$0.0020} \\
 &  & cgcnn & 0.0793{\tiny$\pm$0.0048} & 0.0773{\tiny$\pm$0.0071} \\
 &  & dimenet & 0.0719{\tiny$\pm$0.0019} & 0.0674{\tiny$\pm$0.0026} \\
 &  & schnet & 0.0726{\tiny$\pm$0.0005} & 0.0700{\tiny$\pm$0.0009} \\
 & struct\_monoclinic & alignn & 0.0600{\tiny$\pm$0.0019} & 0.0614{\tiny$\pm$0.0028} \\
 &  & cgcnn & 0.0777{\tiny$\pm$0.0051} & 0.0769{\tiny$\pm$0.0056} \\
 &  & dimenet & 0.0677{\tiny$\pm$0.0058} & 0.0703{\tiny$\pm$0.0062} \\
 &  & schnet & 0.0710{\tiny$\pm$0.0008} & 0.0712{\tiny$\pm$0.0030} \\
 & struct\_orthorhombic & alignn & 0.0578{\tiny$\pm$0.0002} & 0.0608{\tiny$\pm$0.0015} \\
 &  & cgcnn & 0.0724{\tiny$\pm$0.0016} & 0.0803{\tiny$\pm$0.0018} \\
 &  & dimenet & 0.0672{\tiny$\pm$0.0016} & 0.0732{\tiny$\pm$0.0035} \\
 &  & schnet & 0.0672{\tiny$\pm$0.0038} & 0.0737{\tiny$\pm$0.0007} \\
 & struct\_perovskite & alignn & 0.0563{\tiny$\pm$0.0020} & 0.0550{\tiny$\pm$0.0032} \\
 &  & cgcnn & 0.0771{\tiny$\pm$0.0058} & 0.0749{\tiny$\pm$0.0064} \\
 &  & dimenet & 0.0645{\tiny$\pm$0.0018} & 0.0597{\tiny$\pm$0.0012} \\
 &  & schnet & 0.0700{\tiny$\pm$0.0044} & 0.0641{\tiny$\pm$0.0034} \\
 & struct\_tetragonal & alignn & 0.0601{\tiny$\pm$0.0052} & 0.0593{\tiny$\pm$0.0044} \\
 &  & cgcnn & 0.0797{\tiny$\pm$0.0057} & 0.0768{\tiny$\pm$0.0042} \\
 &  & dimenet & 0.0699{\tiny$\pm$0.0019} & 0.0689{\tiny$\pm$0.0036} \\
 &  & schnet & 0.0695{\tiny$\pm$0.0057} & 0.0699{\tiny$\pm$0.0024} \\
 & struct\_triclinic & alignn & 0.0597{\tiny$\pm$0.0028} & 0.0585{\tiny$\pm$0.0020} \\
 &  & cgcnn & 0.0791{\tiny$\pm$0.0067} & 0.0676{\tiny$\pm$0.0039} \\
 &  & dimenet & 0.0644{\tiny$\pm$0.0003} & 0.0581{\tiny$\pm$0.0009} \\
 &  & schnet & 0.0703{\tiny$\pm$0.0041} & 0.0667{\tiny$\pm$0.0045} \\
\addlinespace[2pt]
perovskites & comp\_chalcogenide & alignn & 0.0302{\tiny$\pm$0.0010} & 0.0286{\tiny$\pm$0.0020} \\
 &  & cgcnn & 0.0636{\tiny$\pm$0.0060} & 0.0622{\tiny$\pm$0.0041} \\
 &  & dimenet & 0.0397{\tiny$\pm$0.0057} & 0.0383{\tiny$\pm$0.0049} \\
 &  & schnet & 0.0415{\tiny$\pm$0.0066} & 0.0431{\tiny$\pm$0.0073} \\
 & comp\_halide & alignn & 0.0361{\tiny$\pm$0.0010} & 0.0344{\tiny$\pm$0.0011} \\
 &  & cgcnn & 0.0719{\tiny$\pm$0.0057} & 0.0697{\tiny$\pm$0.0067} \\
 &  & dimenet & 0.0487{\tiny$\pm$0.0085} & 0.0459{\tiny$\pm$0.0076} \\
 &  & schnet & 0.0485{\tiny$\pm$0.0046} & 0.0463{\tiny$\pm$0.0039} \\
 & comp\_nitride & alignn & 0.0318{\tiny$\pm$0.0015} & 0.0316{\tiny$\pm$0.0014} \\
 &  & cgcnn & 0.0674{\tiny$\pm$0.0062} & 0.0654{\tiny$\pm$0.0054} \\
 &  & dimenet & 0.0436{\tiny$\pm$0.0056} & 0.0430{\tiny$\pm$0.0067} \\
 &  & schnet & 0.0411{\tiny$\pm$0.0032} & 0.0404{\tiny$\pm$0.0037} \\
 & comp\_oxide & alignn & 0.0318{\tiny$\pm$0.0015} & 0.0316{\tiny$\pm$0.0015} \\
 &  & cgcnn & 0.0674{\tiny$\pm$0.0062} & 0.0654{\tiny$\pm$0.0054} \\
 &  & dimenet & 0.0440{\tiny$\pm$0.0071} & 0.0437{\tiny$\pm$0.0076} \\
 &  & schnet & 0.0430{\tiny$\pm$0.0038} & 0.0421{\tiny$\pm$0.0045} \\
 & struct\_cubic & alignn & 0.0338{\tiny$\pm$0.0014} & 0.0317{\tiny$\pm$0.0015} \\
 &  & cgcnn & 0.0678{\tiny$\pm$0.0047} & 0.0661{\tiny$\pm$0.0046} \\
 &  & dimenet & 0.0441{\tiny$\pm$0.0082} & 0.0433{\tiny$\pm$0.0076} \\
 &  & schnet & 0.0454{\tiny$\pm$0.0043} & 0.0438{\tiny$\pm$0.0048} \\
 & struct\_high\_spacegroup & alignn & 0.0338{\tiny$\pm$0.0014} & 0.0321{\tiny$\pm$0.0011} \\
 &  & cgcnn & 0.0678{\tiny$\pm$0.0047} & 0.0661{\tiny$\pm$0.0046} \\
 &  & dimenet & 0.0431{\tiny$\pm$0.0060} & 0.0427{\tiny$\pm$0.0062} \\
 &  & schnet & 0.0436{\tiny$\pm$0.0034} & 0.0409{\tiny$\pm$0.0032} \\
 & struct\_orthorhombic & alignn & 0.0333{\tiny$\pm$0.0002} & 0.0351{\tiny$\pm$0.0004} \\
 &  & cgcnn & 0.0682{\tiny$\pm$0.0011} & 0.0689{\tiny$\pm$0.0011} \\
 &  & dimenet & 0.0476{\tiny$\pm$0.0078} & 0.0496{\tiny$\pm$0.0082} \\
 &  & schnet & 0.0483{\tiny$\pm$0.0080} & 0.0500{\tiny$\pm$0.0077} \\
 & struct\_perovskite & alignn & 0.0334{\tiny$\pm$0.0022} & 0.0340{\tiny$\pm$0.0011} \\
 &  & cgcnn & 0.0673{\tiny$\pm$0.0050} & 0.0698{\tiny$\pm$0.0051} \\
 &  & dimenet & 0.0415{\tiny$\pm$0.0038} & 0.0442{\tiny$\pm$0.0041} \\
 &  & schnet & 0.0471{\tiny$\pm$0.0073} & 0.0478{\tiny$\pm$0.0073} \\
 & struct\_tetragonal & alignn & 0.0398{\tiny$\pm$0.0027} & 0.0408{\tiny$\pm$0.0020} \\
 &  & cgcnn & 0.0803{\tiny$\pm$0.0056} & 0.0815{\tiny$\pm$0.0064} \\
 &  & dimenet & 0.0544{\tiny$\pm$0.0074} & 0.0541{\tiny$\pm$0.0079} \\
 &  & schnet & 0.0564{\tiny$\pm$0.0056} & 0.0566{\tiny$\pm$0.0043} \\
\addlinespace[2pt]
phonons & comp\_chalcogenide & alignn & 38.9{\tiny$\pm$8.5} & 36.0{\tiny$\pm$10.3} \\
 &  & cgcnn & 84.5{\tiny$\pm$3.1} & 87.0{\tiny$\pm$4.5} \\
 &  & dimenet & 62.6{\tiny$\pm$13.1} & 64.5{\tiny$\pm$5.9} \\
 &  & schnet & 65.7{\tiny$\pm$13.4} & 67.1{\tiny$\pm$10.6} \\
 & comp\_halide & alignn & 34.5{\tiny$\pm$11.5} & 51.4{\tiny$\pm$13.0} \\
 &  & cgcnn & 79.2{\tiny$\pm$3.3} & 108.6{\tiny$\pm$1.9} \\
 &  & dimenet & 39.7{\tiny$\pm$0.7} & 72.2{\tiny$\pm$7.8} \\
 &  & schnet & 48.0{\tiny$\pm$5.4} & 70.2{\tiny$\pm$6.1} \\
 & comp\_intermetallic & alignn & 36.5{\tiny$\pm$17.6} & 33.5{\tiny$\pm$11.1} \\
 &  & cgcnn & 81.7{\tiny$\pm$9.3} & 84.5{\tiny$\pm$5.2} \\
 &  & dimenet & 82.0{\tiny$\pm$10.8} & 59.1{\tiny$\pm$5.4} \\
 &  & schnet & 79.7{\tiny$\pm$1.6} & 65.0{\tiny$\pm$9.1} \\
 & comp\_nitride & alignn & 27.7{\tiny$\pm$7.0} & 40.0{\tiny$\pm$3.6} \\
 &  & cgcnn & 64.8{\tiny$\pm$12.5} & 68.5{\tiny$\pm$4.7} \\
 &  & dimenet & 57.9{\tiny$\pm$12.1} & 89.0{\tiny$\pm$26.1} \\
 &  & schnet & 52.5{\tiny$\pm$4.2} & 66.5{\tiny$\pm$9.6} \\
 & comp\_oxide & alignn & 44.3{\tiny$\pm$17.2} & 42.3{\tiny$\pm$16.5} \\
 &  & cgcnn & 76.0{\tiny$\pm$7.3} & 96.4{\tiny$\pm$6.8} \\
 &  & dimenet & 59.2{\tiny$\pm$8.2} & 68.7{\tiny$\pm$5.5} \\
 &  & schnet & 61.9{\tiny$\pm$6.6} & 72.1{\tiny$\pm$0.9} \\
 & struct\_cubic & alignn & 37.0{\tiny$\pm$8.8} & 49.7{\tiny$\pm$11.3} \\
 &  & cgcnn & 62.4{\tiny$\pm$2.8} & 96.7{\tiny$\pm$1.7} \\
 &  & dimenet & 42.7{\tiny$\pm$10.4} & 69.5{\tiny$\pm$7.3} \\
 &  & schnet & 51.6{\tiny$\pm$9.9} & 80.3{\tiny$\pm$6.1} \\
 & struct\_hexagonal & alignn & 31.0{\tiny$\pm$3.5} & 32.3{\tiny$\pm$4.4} \\
 &  & cgcnn & 81.4{\tiny$\pm$5.5} & 79.4{\tiny$\pm$6.4} \\
 &  & dimenet & 58.2{\tiny$\pm$7.7} & 55.7{\tiny$\pm$8.8} \\
 &  & schnet & 67.9{\tiny$\pm$2.8} & 63.5{\tiny$\pm$3.2} \\
 & struct\_high\_spacegroup & alignn & 37.0{\tiny$\pm$8.8} & 49.7{\tiny$\pm$11.3} \\
 &  & cgcnn & 62.4{\tiny$\pm$2.8} & 96.7{\tiny$\pm$1.7} \\
 &  & dimenet & 41.1{\tiny$\pm$6.5} & 71.4{\tiny$\pm$10.3} \\
 &  & schnet & 46.7{\tiny$\pm$2.4} & 72.5{\tiny$\pm$5.2} \\
 & struct\_low\_spacegroup & alignn & 40.8{\tiny$\pm$4.7} & 33.1{\tiny$\pm$5.9} \\
 &  & cgcnn & 93.9{\tiny$\pm$3.3} & 77.9{\tiny$\pm$10.7} \\
 &  & dimenet & 68.3{\tiny$\pm$19.9} & 59.4{\tiny$\pm$13.6} \\
 &  & schnet & 60.2{\tiny$\pm$8.6} & 55.3{\tiny$\pm$4.1} \\
 & struct\_monoclinic & alignn & 25.7{\tiny$\pm$5.1} & 28.4{\tiny$\pm$6.0} \\
 &  & cgcnn & 66.0{\tiny$\pm$4.1} & 84.0{\tiny$\pm$6.5} \\
 &  & dimenet & 42.6{\tiny$\pm$9.2} & 54.8{\tiny$\pm$13.3} \\
 &  & schnet & 50.4{\tiny$\pm$6.0} & 46.6{\tiny$\pm$2.4} \\
 & struct\_orthorhombic & alignn & 27.9{\tiny$\pm$2.8} & 32.3{\tiny$\pm$3.7} \\
 &  & cgcnn & 78.9{\tiny$\pm$1.7} & 83.3{\tiny$\pm$1.0} \\
 &  & dimenet & 66.2{\tiny$\pm$6.4} & 72.3{\tiny$\pm$10.3} \\
 &  & schnet & 61.8{\tiny$\pm$5.5} & 63.4{\tiny$\pm$6.0} \\
 & struct\_perovskite & alignn & 33.9{\tiny$\pm$5.7} & 34.3{\tiny$\pm$3.1} \\
 &  & cgcnn & 89.4{\tiny$\pm$16.9} & 80.7{\tiny$\pm$2.1} \\
 &  & dimenet & 78.0{\tiny$\pm$8.7} & 54.6{\tiny$\pm$8.1} \\
 &  & schnet & 58.2{\tiny$\pm$1.7} & 59.1{\tiny$\pm$5.4} \\
 & struct\_tetragonal & alignn & 38.4{\tiny$\pm$11.7} & 35.0{\tiny$\pm$6.1} \\
 &  & cgcnn & 91.9{\tiny$\pm$7.4} & 79.3{\tiny$\pm$3.3} \\
 &  & dimenet & 56.1{\tiny$\pm$14.2} & 69.8{\tiny$\pm$14.4} \\
 &  & schnet & 53.9{\tiny$\pm$5.1} & 65.7{\tiny$\pm$4.1} \\
\end{longtable}}

\paragraph{Per-detector AUROC by architecture and split type.}
Table~\ref{tab:per_arch_split_combined} breaks the per-detector AUROC
down by GNN backbone (left four columns) and by split family (right
two columns). No detector exceeds $0.55$ mean AUROC in any
column except Cosine on the compositional subset ($0.556$); the near chance
ceiling otherwise holds across both architecture and split type.

\begin{table}[t]
\centering
\small
\setlength{\tabcolsep}{4pt}
\caption{\textbf{Per-detector AUROC, stratified by architecture and split type (mean $\pm$ std over three encoder seeds).} Compositional $N{=}104$, structural $N{=}160$ settings. No detector exceeds $0.55$ in any column except Cosine on the compositional subset ($0.556$). \textbf{Bold} is the best value per column among unsupervised detectors. $\uparrow$ indicates higher is better.}
\label{tab:per_arch_split_combined}
\resizebox{\linewidth}{!}{%
\begin{tabular}{lrrrrrr}
\toprule
 & \multicolumn{4}{c}{Architecture} & \multicolumn{2}{c}{Split type} \\
\cmidrule(lr){2-5}\cmidrule(lr){6-7}
Detector & Alignn $\uparrow$ & Cgcnn $\uparrow$ & Schnet $\uparrow$ & Dimenet $\uparrow$ & Comp.\ $\uparrow$ & Struct.\ $\uparrow$ \\
\midrule
KNN-k20 & 0.502 {\tiny$\pm$0.002} & 0.518 {\tiny$\pm$0.007} & 0.509 {\tiny$\pm$0.003} & 0.519 {\tiny$\pm$0.002} & 0.529 {\tiny$\pm$0.002} & 0.501 {\tiny$\pm$0.001} \\
KNN-k50 & 0.497 {\tiny$\pm$0.002} & 0.514 {\tiny$\pm$0.008} & 0.506 {\tiny$\pm$0.003} & 0.517 {\tiny$\pm$0.001} & 0.523 {\tiny$\pm$0.003} & 0.499 {\tiny$\pm$0.002} \\
Norm & 0.512 {\tiny$\pm$0.001} & 0.503 {\tiny$\pm$0.002} & 0.491 {\tiny$\pm$0.001} & 0.494 {\tiny$\pm$0.001} & 0.502 {\tiny$\pm$0.003} & 0.498 {\tiny$\pm$0.001} \\
Cosine & 0.530 {\tiny$\pm$0.001} & \textbf{0.535 {\tiny$\pm$0.003}} & \textbf{0.529 {\tiny$\pm$0.001}} & \textbf{0.530 {\tiny$\pm$0.002}} & \textbf{0.556 {\tiny$\pm$0.001}} & \textbf{0.515 {\tiny$\pm$0.001}} \\
LOF & 0.536 {\tiny$\pm$0.002} & 0.530 {\tiny$\pm$0.003} & 0.515 {\tiny$\pm$0.002} & 0.522 {\tiny$\pm$0.001} & 0.545 {\tiny$\pm$0.002} & 0.513 {\tiny$\pm$0.001} \\
ASH+KNN & 0.532 {\tiny$\pm$0.001} & 0.517 {\tiny$\pm$0.004} & 0.506 {\tiny$\pm$0.003} & 0.510 {\tiny$\pm$0.008} & 0.536 {\tiny$\pm$0.001} & 0.504 {\tiny$\pm$0.003} \\
MDS-BIC & 0.532 {\tiny$\pm$0.002} & 0.520 {\tiny$\pm$0.005} & 0.498 {\tiny$\pm$0.000} & 0.505 {\tiny$\pm$0.001} & 0.530 {\tiny$\pm$0.002} & 0.503 {\tiny$\pm$0.000} \\
KDE & \textbf{0.539 {\tiny$\pm$0.001}} & 0.532 {\tiny$\pm$0.002} & 0.498 {\tiny$\pm$0.001} & 0.504 {\tiny$\pm$0.001} & 0.537 {\tiny$\pm$0.001} & 0.506 {\tiny$\pm$0.001} \\
Flow & 0.536 {\tiny$\pm$0.004} & 0.524 {\tiny$\pm$0.002} & 0.510 {\tiny$\pm$0.011} & 0.509 {\tiny$\pm$0.001} & 0.543 {\tiny$\pm$0.007} & 0.505 {\tiny$\pm$0.001} \\
kPCA & 0.512 {\tiny$\pm$0.001} & 0.516 {\tiny$\pm$0.006} & 0.507 {\tiny$\pm$0.003} & 0.514 {\tiny$\pm$0.001} & 0.529 {\tiny$\pm$0.003} & 0.502 {\tiny$\pm$0.001} \\
ReAct+MDS & 0.525 {\tiny$\pm$0.003} & 0.520 {\tiny$\pm$0.004} & 0.499 {\tiny$\pm$0.002} & 0.505 {\tiny$\pm$0.001} & 0.529 {\tiny$\pm$0.002} & 0.501 {\tiny$\pm$0.000} \\
IForest & 0.492 {\tiny$\pm$0.001} & 0.508 {\tiny$\pm$0.006} & 0.501 {\tiny$\pm$0.003} & 0.508 {\tiny$\pm$0.002} & 0.509 {\tiny$\pm$0.003} & 0.498 {\tiny$\pm$0.001} \\
ADERH & 0.503 {\tiny$\pm$0.005} & 0.500 {\tiny$\pm$0.005} & 0.492 {\tiny$\pm$0.000} & 0.495 {\tiny$\pm$0.001} & 0.500 {\tiny$\pm$0.004} & 0.495 {\tiny$\pm$0.002} \\
OCSVM & 0.488 {\tiny$\pm$0.000} & 0.506 {\tiny$\pm$0.004} & 0.496 {\tiny$\pm$0.002} & 0.494 {\tiny$\pm$0.002} & 0.492 {\tiny$\pm$0.004} & 0.498 {\tiny$\pm$0.002} \\
ViM & 0.530 {\tiny$\pm$0.003} & 0.514 {\tiny$\pm$0.006} & 0.493 {\tiny$\pm$0.001} & 0.499 {\tiny$\pm$0.001} & 0.521 {\tiny$\pm$0.002} & 0.501 {\tiny$\pm$0.001} \\
\bottomrule
\end{tabular}}
\end{table}

\paragraph{Mean AUROC by detector family and dataset.}
Table~\ref{tab:family_by_dataset} aggregates the same per-setting AUROCs
along the dataset axis instead of the architecture / split type axis.
Two patterns emerge. First, no unsupervised family reaches mean AUROC
$\geq 0.55$ on any individual dataset: the column maxima top out at
$0.531$ (Distance, dielectric) and $0.534$ (Density, log\_gvrh). This shows that the ``no detector mechanism escapes the near chance ceiling''
claim from Section~\ref{sec:eval_uniform} of the main paper holds at the
family\,$\times$\,dataset granularity, not just in aggregate. Second,
the cross-dataset spread tracks the bottleneck dichotomy directly. On
the method-bottlenecked datasets (dielectric, log\_gvrh) the
unsupervised families sit at $0.49$--$0.53$ while the
supervised ceiling jumps to $0.60$, leaving a $\sim$$0.07$
gap that captures the OOD signal unsupervised detectors fail to extract.
On the encoder-bottlenecked datasets (log\_kvrh, perovskites, phonons)
every family, including supervised, sits within $0.008$ of
chance: the OOD signal is not in the embeddings for any method,
unsupervised or otherwise, to recover.

\begin{table}[t]
\centering
\small
\setlength{\tabcolsep}{4pt}
\caption{\textbf{Mean AUROC by detector family and dataset (mean $\pm$ std over three encoder seeds).} Each row aggregates the per-setting AUROCs of every detector in that family. The supervised row is a diagnostic ceiling, excluded from rankings. No unsupervised family reaches $\geq 0.55$ on any dataset. \textbf{Bold} is the best value per column among unsupervised detectors. $\uparrow$ indicates higher is better.}
\label{tab:family_by_dataset}
\resizebox{\linewidth}{!}{%
\begin{tabular}{lrrrrrr}
\toprule
Family & All $\uparrow$ & dielectric $\uparrow$ & log\_gvrh $\uparrow$ & log\_kvrh $\uparrow$ & perovskites $\uparrow$ & phonons $\uparrow$ \\
\midrule
Distance & \textbf{0.516 {\tiny$\pm$0.000}} & \textbf{0.531 {\tiny$\pm$0.001}} & 0.530 {\tiny$\pm$0.001} & \textbf{0.506 {\tiny$\pm$0.000}} & \textbf{0.501 {\tiny$\pm$0.001}} & 0.502 {\tiny$\pm$0.003} \\
Density & 0.515 {\tiny$\pm$0.001} & 0.530 {\tiny$\pm$0.002} & \textbf{0.534 {\tiny$\pm$0.001}} & 0.505 {\tiny$\pm$0.001} & 0.500 {\tiny$\pm$0.001} & 0.499 {\tiny$\pm$0.003} \\
Isolation & 0.500 {\tiny$\pm$0.002} & 0.494 {\tiny$\pm$0.004} & 0.498 {\tiny$\pm$0.003} & 0.502 {\tiny$\pm$0.001} & 0.497 {\tiny$\pm$0.001} & \textbf{0.507 {\tiny$\pm$0.003}} \\
Boundary / Hybrid & 0.502 {\tiny$\pm$0.001} & 0.501 {\tiny$\pm$0.003} & 0.507 {\tiny$\pm$0.002} & 0.500 {\tiny$\pm$0.001} & 0.497 {\tiny$\pm$0.001} & 0.505 {\tiny$\pm$0.004} \\
\midrule
Supervised & 0.543 {\tiny$\pm$0.001} & 0.600 {\tiny$\pm$0.003} & 0.603 {\tiny$\pm$0.002} & 0.492 {\tiny$\pm$0.003} & 0.499 {\tiny$\pm$0.004} & 0.496 {\tiny$\pm$0.008} \\
\bottomrule
\end{tabular}}
\end{table}

\paragraph{FPR95.}
Table~\ref{tab:fpr95_summary} reports FPR95 (false-positive rate at TPR
$=$ 95\%; lower is better) for each detector and each dataset. Mean FPR95
$\geq 0.93$ in every column mirrors the near chance AUROC ceiling.

\begin{table}[t]
\centering
\small
\setlength{\tabcolsep}{3pt}
\caption{\textbf{FPR95 per detector and per dataset (mean $\pm$ std over three encoder seeds).} Lower is better. Covers 14 of 15 unsupervised detectors (Flow FPR95 is deferred to v2). KNN-k10 is the default $k{=}10$ KNN from the per-setting evaluation outputs, distinct from the $k{=}20$/$k{=}50$ variants in Table~\ref{tab:per_detector_summary}. \textbf{Bold} is the best value per column among unsupervised detectors. $\downarrow$ indicates lower is better.}
\label{tab:fpr95_summary}
\resizebox{\linewidth}{!}{%
\begin{tabular}{lrrrrrr}
\toprule
Detector & All $\downarrow$ & dielectric $\downarrow$ & log\_gvrh $\downarrow$ & log\_kvrh $\downarrow$ & perovskites $\downarrow$ & phonons $\downarrow$ \\
\midrule
MDS-BIC & 0.935 {\tiny$\pm$0.002} & 0.926 {\tiny$\pm$0.004} & 0.917 {\tiny$\pm$0.003} & 0.946 {\tiny$\pm$0.001} & 0.953 {\tiny$\pm$0.001} & 0.939 {\tiny$\pm$0.007} \\
KNN-k10 & 0.942 {\tiny$\pm$0.002} & 0.935 {\tiny$\pm$0.005} & 0.937 {\tiny$\pm$0.004} & 0.945 {\tiny$\pm$0.002} & 0.949 {\tiny$\pm$0.002} & 0.946 {\tiny$\pm$0.005} \\
kPCA & 0.938 {\tiny$\pm$0.002} & 0.926 {\tiny$\pm$0.004} & 0.929 {\tiny$\pm$0.005} & \textbf{0.945 {\tiny$\pm$0.002}} & 0.950 {\tiny$\pm$0.001} & 0.944 {\tiny$\pm$0.006} \\
Norm & 0.947 {\tiny$\pm$0.002} & 0.953 {\tiny$\pm$0.003} & 0.939 {\tiny$\pm$0.002} & 0.948 {\tiny$\pm$0.001} & 0.951 {\tiny$\pm$0.001} & 0.946 {\tiny$\pm$0.006} \\
LOF & 0.928 {\tiny$\pm$0.001} & 0.910 {\tiny$\pm$0.005} & \textbf{0.894 {\tiny$\pm$0.004}} & 0.945 {\tiny$\pm$0.002} & 0.951 {\tiny$\pm$0.003} & 0.953 {\tiny$\pm$0.006} \\
OCSVM & 0.946 {\tiny$\pm$0.002} & 0.943 {\tiny$\pm$0.004} & 0.945 {\tiny$\pm$0.005} & 0.948 {\tiny$\pm$0.000} & 0.950 {\tiny$\pm$0.002} & 0.944 {\tiny$\pm$0.004} \\
IForest & 0.945 {\tiny$\pm$0.003} & 0.947 {\tiny$\pm$0.005} & 0.942 {\tiny$\pm$0.006} & 0.947 {\tiny$\pm$0.001} & 0.950 {\tiny$\pm$0.003} & 0.937 {\tiny$\pm$0.010} \\
ADERH & 0.947 {\tiny$\pm$0.002} & 0.955 {\tiny$\pm$0.002} & 0.943 {\tiny$\pm$0.004} & 0.948 {\tiny$\pm$0.002} & 0.952 {\tiny$\pm$0.001} & 0.940 {\tiny$\pm$0.007} \\
KDE & 0.935 {\tiny$\pm$0.000} & 0.929 {\tiny$\pm$0.003} & 0.914 {\tiny$\pm$0.004} & 0.948 {\tiny$\pm$0.001} & 0.951 {\tiny$\pm$0.003} & 0.939 {\tiny$\pm$0.003} \\
ViM & 0.938 {\tiny$\pm$0.002} & 0.936 {\tiny$\pm$0.002} & 0.923 {\tiny$\pm$0.003} & 0.947 {\tiny$\pm$0.000} & 0.954 {\tiny$\pm$0.002} & \textbf{0.937 {\tiny$\pm$0.008}} \\
Cosine & \textbf{0.927 {\tiny$\pm$0.001}} & \textbf{0.900 {\tiny$\pm$0.003}} & 0.900 {\tiny$\pm$0.002} & 0.946 {\tiny$\pm$0.003} & 0.952 {\tiny$\pm$0.002} & 0.949 {\tiny$\pm$0.004} \\
ReAct+MDS & 0.936 {\tiny$\pm$0.001} & 0.928 {\tiny$\pm$0.004} & 0.920 {\tiny$\pm$0.003} & 0.946 {\tiny$\pm$0.001} & 0.953 {\tiny$\pm$0.002} & 0.940 {\tiny$\pm$0.004} \\
ASH+KNN & 0.933 {\tiny$\pm$0.002} & 0.918 {\tiny$\pm$0.002} & 0.917 {\tiny$\pm$0.002} & 0.945 {\tiny$\pm$0.003} & \textbf{0.949 {\tiny$\pm$0.003}} & 0.943 {\tiny$\pm$0.007} \\
KNN-k20-norm & 0.940 {\tiny$\pm$0.002} & 0.933 {\tiny$\pm$0.004} & 0.931 {\tiny$\pm$0.004} & 0.945 {\tiny$\pm$0.001} & 0.951 {\tiny$\pm$0.001} & 0.944 {\tiny$\pm$0.005} \\
KNN-k50-norm & 0.941 {\tiny$\pm$0.002} & 0.938 {\tiny$\pm$0.003} & 0.935 {\tiny$\pm$0.004} & 0.945 {\tiny$\pm$0.001} & 0.949 {\tiny$\pm$0.002} & 0.940 {\tiny$\pm$0.006} \\
\bottomrule
\end{tabular}}
\end{table}

\begin{table}[t]
\centering
\small
\setlength{\tabcolsep}{4pt}
\caption{\textbf{Per-setting permutation test of the linear probe ($B{=}100$, same 5-fold protocol), per encoder seed.} $p=(1+\#\{\mathrm{null}\ge\mathrm{real}\})/(B+1)$. Rejection rates are mean $\pm$ std across seeds; per-seed counts given as $c_{42}/c_{123}/c_{456}$ out of $N$. The linear probe column is from the canonical probe pipeline. Under the null, $\sim$1\% of settings reach $p<0.01$ by chance.}
\label{tab:permutation_test}
\begin{tabular}{lrcrrl}
\toprule
Dataset & $N$ & Mean linear probe & \% $p<0.01$ & \% $p<0.05$ & Bottleneck \\
\midrule
\rowcolor{yellow!20}
dielectric & 56 & 0.599 $\pm$ 0.001 & 28.6 $\pm$ 0.0 (16/56, 16/56, 16/56) & 31.5 $\pm$ 1.0 (18/56, 17/56, 18/56) & Method \\
\rowcolor{yellow!20}
log\_gvrh & 60 & 0.600 $\pm$ 0.002 & 26.7 $\pm$ 0.0 (16/60, 16/60, 16/60) & 26.7 $\pm$ 0.0 (16/60, 16/60, 16/60) & Method \\
\rowcolor{blue!12}
log\_kvrh & 60 & 0.494 $\pm$ 0.004 & 0.0 $\pm$ 0.0 (0/60, 0/60, 0/60) & 4.4 $\pm$ 4.8 (1/60, 1/60, 6/60) & Encoder \\
\rowcolor{blue!12}
perovskites & 36 & 0.500 $\pm$ 0.005 & 0.0 $\pm$ 0.0 (0/36, 0/36, 0/36) & 2.8 $\pm$ 2.8 (0/36, 1/36, 2/36) & Encoder \\
\rowcolor{blue!12}
phonons & 52 & 0.499 $\pm$ 0.009 & 0.6 $\pm$ 1.1 (0/45, 0/52, 1/52) & 1.9 $\pm$ 1.9 (0/45, 1/52, 2/52) & Encoder \\
\bottomrule
\end{tabular}
\end{table}

\begin{table}[h]
\centering
\small
\caption{\textbf{Full per-detector OLS of AUROC against $\log_{10}(\mathrm{MAE}_{\mathrm{OOD}}/\mathrm{MAE}_{\mathrm{ID}})$} (mean $\pm$ std over three encoder seeds), on all $N_{all}=264$ settings and the degraded subset. \textbf{Bold} is the best value per column. $\uparrow$ indicates higher is better.}
\label{tab:regression_auroc_mae_gap}
\begin{tabular}{lrrrr}
\toprule
Detector & $\beta$ (all) $\uparrow$ & $R^2$ (all) $\uparrow$ & $\beta$ (degraded) $\uparrow$ & $R^2$ (degraded) $\uparrow$ \\
\midrule
KNN-k20-norm & +0.215 $\pm$ 0.020 & 0.22 $\pm$ 0.01 & +0.295 $\pm$ 0.009 & 0.20 $\pm$ 0.01 \\
KNN-k50-norm & +0.191 $\pm$ 0.019 & 0.18 $\pm$ 0.01 & +0.243 $\pm$ 0.013 & 0.15 $\pm$ 0.01 \\
Norm & +0.069 $\pm$ 0.018 & 0.04 $\pm$ 0.02 & +0.042 $\pm$ 0.016 & 0.01 $\pm$ 0.00 \\
Cosine & +0.311 $\pm$ 0.027 & \textbf{0.26 $\pm$ 0.02} & +0.579 $\pm$ 0.024 & \textbf{0.43 $\pm$ 0.02} \\
LOF & +0.259 $\pm$ 0.042 & 0.22 $\pm$ 0.06 & +0.466 $\pm$ 0.027 & 0.36 $\pm$ 0.03 \\
ASH+KNN & +0.215 $\pm$ 0.032 & 0.21 $\pm$ 0.04 & +0.347 $\pm$ 0.053 & 0.29 $\pm$ 0.08 \\
MDS-BIC & +0.218 $\pm$ 0.027 & 0.20 $\pm$ 0.03 & +0.297 $\pm$ 0.025 & 0.18 $\pm$ 0.03 \\
KDE & +0.230 $\pm$ 0.019 & 0.19 $\pm$ 0.02 & +0.362 $\pm$ 0.010 & 0.23 $\pm$ 0.01 \\
Flow & +0.259 $\pm$ 0.018 & 0.23 $\pm$ 0.04 & +0.416 $\pm$ 0.029 & 0.28 $\pm$ 0.02 \\
kPCA & +0.226 $\pm$ 0.030 & 0.24 $\pm$ 0.04 & +0.323 $\pm$ 0.032 & 0.24 $\pm$ 0.04 \\
ReAct+MDS & +0.219 $\pm$ 0.025 & 0.21 $\pm$ 0.03 & +0.304 $\pm$ 0.018 & 0.20 $\pm$ 0.02 \\
IForest & +0.157 $\pm$ 0.030 & 0.17 $\pm$ 0.05 & +0.178 $\pm$ 0.031 & 0.12 $\pm$ 0.04 \\
ADERH & +0.069 $\pm$ 0.028 & 0.06 $\pm$ 0.04 & +0.004 $\pm$ 0.039 & 0.01 $\pm$ 0.01 \\
OCSVM & +0.067 $\pm$ 0.030 & 0.03 $\pm$ 0.03 & +0.073 $\pm$ 0.043 & 0.02 $\pm$ 0.02 \\
ViM & +0.177 $\pm$ 0.024 & 0.16 $\pm$ 0.03 & +0.221 $\pm$ 0.032 & 0.13 $\pm$ 0.04 \\
Linear probe & +0.309 $\pm$ 0.043 & 0.12 $\pm$ 0.03 & +0.736 $\pm$ 0.039 & 0.38 $\pm$ 0.04 \\
MLP probe & \textbf{+0.348 $\pm$ 0.046} & 0.15 $\pm$ 0.03 & \textbf{+0.782 $\pm$ 0.049} & 0.40 $\pm$ 0.04 \\
\bottomrule
\end{tabular}
\end{table}

\FloatBarrier

\section{Extended analysis}
\label{app:extended}

\subsection{Detector-level failure modes}
\label{app:detector_insights}

The aggregate near chance performance shown in
Tables~\ref{tab:per_detector_summary} and~\ref{tab:bottleneck_summary}
averages over different detector behaviour. Three specific failure
modes are worth flagging because they are not generic ``every detector
underperformed'' and may help future work.

\paragraph{Embedding norm achieves chance and rarely tracks shift.}
The plain embedding norm detector (mean AUROC $0.500$, median $0.500$)
is no better than a coin flip on the unsupervised side. Its regression
slope against $\log(\text{OOD MAE} / \text{ID MAE})$ is among the smallest of
any detector (Table~\ref{tab:regression_auroc_mae_gap}: degraded
$\beta = 0.042$, $R^2 = 0.01$). One possible explanation is that
embedding norms in supervised graph regression are mostly controlled
by the regression head's normalisation, not by distribution shift
signal: norm size correlates with target size rather than with how
familiar the input is. Logit-based scores (e.g., MSP, energy) have no direct analogue for a
scalar regression head.

\paragraph{ASH+KNN does not improve on plain KNN.}
ASH (activation shaping) was designed to suppress non-essential
activations and improve detection on classification networks. On our
regression encoders ASH+KNN ($\mu = 0.516$, win count $31$) matches
plain KNN with the same $k = 20$ neighbour count ($\mu = 0.512$, win
count $13$) and underperforms the simple Cosine baseline ($\mu =
0.531$, win count $55$). ASH's pruning mask is calibrated against an
activation distribution shaped by softmax-driven sparsity; the dense
regression embeddings here have no such structure for ASH to use.

\paragraph{ViM's virtual-logit term is undefined for regression.}
ViM combines a logit-space score with a feature-space residual norm,
relying on a classification head that produces logits. Our GNNs
produce a single scalar output without a class manifold, so we score
only the embedding-subspace residual: this loses the logit
cross coupling responsible for much of ViM's reported gain on
classification benchmarks. Performance is near chance ($\mu = 0.509$,
above-$0.55$ rate $8.5\%$), and adapting ViM style methods to
regression encoders is an open direction.

\subsection{Robustness analyses}
\label{app:robustness}

We address three possible alternative explanations for the bottleneck
dichotomy. The first is summarised in the main paper; the latter two
are new in this appendix.

\paragraph{Probe non-linearity is not the bottleneck.}
The linear and MLP probes agree to within mean $|\Delta| = 0.023$ across
all 264 settings; see main-paper Figure~\ref{fig:bottleneck_dichotomy}A.

\paragraph{Excluded settings.}
Seven CGCNN/phonons encoders collapse to constant embeddings under seed 42
(variance below $10^{-10}$). All seven train normally under seeds 123 and 456;
we retain these settings and average over the two valid seeds, so no setting
is excluded for collapse.

\begin{table}[h]
\centering
\caption{\textbf{Linear-probe AUROC by dataset and architecture} (mean $\pm$ std over three seeds).
All 20 cells fall on the same side of the 0.55 threshold as the dataset-level label. \textbf{Bold} is the best value per row. $\uparrow$ indicates higher is better.}
\label{tab:per_arch_probe}
\resizebox{\linewidth}{!}{%
\begin{tabular}{lcccccl}
\toprule
Dataset & ALIGNN $\uparrow$ & CGCNN $\uparrow$ & SchNet $\uparrow$ & DimeNet++ $\uparrow$ & Spread & Label \\
\midrule
dielectric  & \textbf{0.612 $\pm$ 0.003} & 0.599 $\pm$ 0.002 & 0.596 $\pm$ 0.002 & 0.591 $\pm$ 0.004 & 0.022 & Method \\
log\_gvrh   & \textbf{0.612 $\pm$ 0.002} & 0.608 $\pm$ 0.001 & 0.595 $\pm$ 0.003 & 0.585 $\pm$ 0.003 & 0.027 & Method \\
log\_kvrh   & 0.488 $\pm$ 0.004 & \textbf{0.502 $\pm$ 0.010} & 0.496 $\pm$ 0.006 & 0.491 $\pm$ 0.007 & 0.014 & Encoder \\
perovskites & 0.495 $\pm$ 0.014 & \textbf{0.504 $\pm$ 0.006} & \textbf{0.504 $\pm$ 0.002} & 0.498 $\pm$ 0.005 & 0.009 & Encoder \\
phonons     & \textbf{0.521 $\pm$ 0.008} & 0.498 $\pm$ 0.019 & 0.491 $\pm$ 0.011 & 0.485 $\pm$ 0.020 & 0.036 & Encoder \\
\bottomrule
\end{tabular}}
\end{table}

\paragraph{Per-setting seed stability.}
Beyond the dataset-level labels (Table~\ref{tab:per_arch_probe}), the per-setting
linear probe AUROC is stable across the three encoder seeds. At the dataset level,
seed-to-seed standard deviations are $0.001$--$0.009$, an order of magnitude below
the $0.101$--$0.105$ separation between the two regimes in mean probe AUROC. At the
individual-setting level the seed std is larger, ranging from $0.000$ to $0.092$,
which is why a minority of settings cross the $0.55$ cutoff on a single seed. Of
the $66$ settings per architecture, the number whose probe stays below $0.55$ on
\emph{every} valid seed is $52$ (ALIGNN), $49$ (CGCNN), $51$ (SchNet) and $53$
(DimeNet++); only $23$ settings cross the $0.55$ cutoff on exactly one seed and
$2$ on exactly two seeds (never flipping a dataset label);
these one-seed crossings are concentrated in phonons, the smallest task ($14$ of
$23$), with the rest in log\_kvrh ($4$), dielectric ($3$) and perovskites ($2$).
The single highest probe AUROC in any (setting, seed) cell is $0.999$.

\subsection{Risk-coverage analysis}
\label{app:risk_coverage}

AUROC measures whether a detector ranks ID points above OOD points,
but it does not directly answer ``does the detector reduce downstream
prediction error if we abstain on its highest anomaly samples?''
\citet{jaeger2023call} argue this is the deployment relevant question,
and propose AURC (Area Under the Risk-Coverage curve, lower is better)
and SPE (Selective Prediction Effectiveness; SPE $= 1$ matches an
oracle abstaining on worst error samples first, SPE $= 0$ matches
random abstention, SPE $< 0$ is worse than random). Risk coverage adds
to the AUROC story rather than replacing it: even when AUROC sits near
chance on average, every detector reported in
Table~\ref{tab:aurc_spe_summary} achieves positive median SPE. This means that their
score ordering correlates non-trivially with prediction error on a
majority of settings. The weakest is the plain embedding-norm baseline, with the lowest mean SPE ($+0.14$, $69\%$ of settings
SPE-positive), matching its near chance AUROC profile.

\begin{table}[t]
\centering
\small
\setlength{\tabcolsep}{4pt}
\caption{\textbf{Risk-coverage summary per detector.} Per-setting AURC (area under the risk-coverage curve, lower is better) and SPE (selective-prediction effectiveness; SPE = 1 matches an oracle abstaining on worst error samples first, SPE = 0 matches random abstention, SPE $<$ 0 is worse than random) aggregated across $N = 257$ seed 42 settings. Unlike the other appendix tables, these values are computed on the seed 42 encoder run only: the risk-coverage curves require per-sample detector scores, which are not stored per seed, so a 3-seed mean $\pm$ std is deferred to the next benchmark version. The eight detectors with directly comparable score direction conventions (Mahalanobis variants, KNN $k{=}10$, LOF, kPCA, OCSVM, IForest, Norm, normalising flow density) are reported. Here KNN $k{=}10$ is the default $k{=}10$ KNN from the per-setting evaluation outputs, distinct from the $k{=}20$/$k{=}50$ variants in Table~\ref{tab:per_detector_summary}. \textbf{Bold} is the best value per column. $\downarrow$ indicates lower is better; $\uparrow$ indicates higher is better.}
\label{tab:aurc_spe_summary}
\begin{tabular}{lrrrrrr}
\toprule
Detector & $N$ & Mean AURC $\downarrow$ & Median AURC $\downarrow$ & Mean SPE $\uparrow$ & Median SPE $\uparrow$ & \% SPE $>$ 0 $\uparrow$ \\
\midrule
MDS-BIC & 257 & 6.936 & 0.088 & +0.340 & +0.349 & 94.2\% \\
KNN k=10 & 257 & 7.571 & 0.084 & +0.363 & +0.348 & 93.4\% \\
LOF & 257 & 6.970 & \textbf{0.082} & +0.347 & +0.306 & 98.8\% \\
kPCA & 257 & 6.700 & 0.085 & \textbf{+0.372} & \textbf{+0.374} & 99.2\% \\
OCSVM & 257 & 7.212 & 0.082 & +0.317 & +0.289 & \textbf{99.6\%} \\
IForest & 257 & \textbf{6.591} & 0.086 & +0.354 & +0.333 & 99.2\% \\
Norm & 257 & 8.124 & 0.096 & +0.136 & +0.148 & 69.3\% \\
Flow & 257 & 7.124 & 0.088 & +0.291 & +0.318 & 86.4\% \\
\bottomrule
\end{tabular}
\end{table}

\subsection{Omee distribution-shift cross-check}
\label{app:omee}

To check that the near chance ceiling is not specific to our split
construction, we re-run the two detectors with directly comparable
implementations (Mahalanobis and KNN) on three alternative split
protocols defined by the Omee suite~\citep{omee2024structurebased}:
LOCO (leave-one-cluster-out), SparseXcluster (sparse covariate
clusters), and SparseYcluster (sparse target clusters). Mean AUROC
stays within about $0.05$ of chance on every (task, split protocol)
combination, matching the near chance story under physics motivated
splits (Table~\ref{tab:omee_cross_check}).

\begin{table}[t]
\centering
\small
\setlength{\tabcolsep}{4pt}
\caption{\textbf{Omee distribution-shift cross-check.} The Omee suite (\citet{omee2024structurebased}) defines three alternative split protocols: \textbf{LOCO} (leave-one-cluster-out), \textbf{SparseXcluster} (sparse covariate clusters), and \textbf{SparseYcluster} (sparse target clusters). We rerun the two detectors with directly comparable implementations (Mahalanobis and KNN) on these splits using the same ALIGNN and CGCNN embeddings as the main paper, 50 folds per split. Mean AUROC remains within $\sim$0.05 of chance, consistent with the near chance story under physics motivated splits. $\downarrow$ indicates lower is better. $\uparrow$ indicates higher is better.}
\label{tab:omee_cross_check}
\begin{tabular}{llrrrrr}
\toprule
Dataset & Split & ID MAE $\downarrow$ & OOD MAE $\downarrow$ & log gap & Maha AUROC $\uparrow$ & KNN AUROC $\uparrow$ \\
\midrule
dielectric & LOCO & 0.429 & 0.408 & -0.021 & 0.503 & 0.501 \\
 & SparseXcluster & 0.469 & 0.537 & +0.068 & 0.507 & 0.519 \\
 & SparseYcluster & 0.463 & 0.685 & +0.222 & 0.502 & 0.509 \\
\addlinespace[2pt]
log\_gvrh & LOCO & 0.089 & 0.089 & -0.001 & 0.500 & 0.498 \\
 & SparseXcluster & 0.090 & 0.091 & +0.001 & 0.544 & 0.547 \\
 & SparseYcluster & 0.090 & 0.081 & -0.009 & 0.528 & 0.523 \\
\bottomrule
\end{tabular}
\end{table}

\subsection{Where this lands in the OOD literature}
\label{app:prior_findings}

\paragraph{Consistency with image classification OOD literature.}
The image classification OOD detection literature has converged on the
finding that no single detector dominates across benchmarks; the
relative ranking of detectors is sensitive to the choice of dataset,
encoder, and shift type~\citep{yang2022openood, guilleescuret2024broad}.
Our finding that no detector exceeds $0.55$ mean AUROC under any
architecture (Table~\ref{tab:per_arch_split_combined}) is the
regression domain analogue: not only does no detector dominate, but
the entire family fails to meaningfully exceed chance on a substantial
fraction of settings. 

\paragraph{Disparity with vision OOD benchmarks.}
On image classification benchmarks (CIFAR-10, ImageNet), post-hoc
detectors commonly reach AUROC above $0.85$, particularly on far-OOD
splits~\citep{yang2022openood}. On our regression encoders, the best
unsupervised detector averages $0.531$, and even the supervised probes
average only about $0.60$ on the method-bottlenecked datasets
(Table~\ref{tab:bottleneck_summary}), although individual settings
approach $1.0$. This gap is what motivates the bottleneck dichotomy.

\paragraph{Relation to GeSS.} GeSS~\citep{zou2024gess} benchmarks OOD
\emph{generalization} of geometric deep learning models across particle
physics, biochemistry, and materials science, and finds that no single
learning algorithm is best across shift types. It does not evaluate
post-hoc OOD detection. MOOD is complementary: it asks whether a shift
can be \emph{detected} from frozen embeddings, rather than whether a
model can generalize across it, and adds a quantitative bottleneck
classification that separates detector failures from encoder failures.

\FloatBarrier

\section{Chemistry-aware analysis of bottleneck regimes}
\label{app:chemistry}

The dataset-level bottleneck classification in
\S\ref{sec:eval_dichotomy} treats each MatBench task as an opaque
unit. This appendix asks why the partition falls where it does. We
argue that the dichotomy is not arbitrary: it tracks the physical
drivers of each target property and how well those drivers line up
with our split axes. Datasets land in the method-bottleneck regime
when the GNN must encode property relevant variation that also
defines the OOD shift, and in the encoder-bottleneck regime when the
property is governed by aggregate quantities the network can predict
accurately while throwing away the OOD defining structure.

\subsection{Property physics and split-axis alignment}
\label{app:chemistry_physics}

Table~\ref{tab:property_physics} summarizes, for each MatBench task,
the dominant physical drivers of the target property and which of our
split axes (compositional family vs.\ structural symmetry) we expect
to be property relevant, meaning a shift along that axis moves the
input through the subspace the GNN must encode in order to predict
accurately.

\begin{table}[h]
  \centering
  \small
  \caption{\textbf{Physical drivers of MatBench targets and split-axis alignment.}
  Compositional axes shift element families (oxide, halide, etc.); structural axes shift lattice symmetry (cubic, low-spacegroup, etc.). A shift is property relevant if the property's physical drivers are sensitive to that axis.}
  \label{tab:property_physics}
  \setlength{\tabcolsep}{4pt}
  \begin{tabular}{@{}l l p{5.6cm} c c@{}}
    \toprule
    Dataset & Target & Dominant drivers & Comp.\ rel. & Struct.\ rel. \\
    \midrule
    dielectric & refractive index $n$ & electronic polarizability, bandgap~\citep{shannon1993dielectric,petousis2017dielectric} & \checkmark & partial \\
    log\_gvrh & $\log G_{\mathrm{VRH}}$ (shear) & bond directionality, anisotropy~\citep{deJong2015elastic} & partial & \checkmark \\
    \midrule
    log\_kvrh & $\log K_{\mathrm{VRH}}$ (bulk) & mean bond stiffness, density~\citep{deJong2015elastic} & weak & weak \\
    perovskites & formation energy & cohesive energy within ABX$_3$~\citep{castelli2012perovskite} & weak & weak \\
    phonons & highest optical phonon freq. & reduced mass, local force constants~\citep{petretto2018phonons} & weak & weak \\
    \bottomrule
  \end{tabular}
\end{table}

The first two rows are method-bottlenecked in our results; the last
three rows are encoder-bottlenecked. The pattern matches the alignment
column: when at least one split axis is property relevant, the
embedding is forced to keep that information, and a supervised probe
can recover it. We expand on the physics below.

\paragraph{Method-bottlenecked: dielectric and log\_gvrh.}
The static refractive index of a crystal is mostly set by electronic
polarizability and the bandgap, both of which depend strongly on
which elements are present and their bonding character: oxides,
chalcogenides, and halides occupy distinct regions in the $(n,
E_{\mathrm{gap}})$
plane~\citep{shannon1993dielectric,petousis2017dielectric}. A
compositional family shift is therefore a shift in the property's
physical inputs, and the GNN must encode element family signal to
achieve low MAE. Probes on these embeddings recover that signal.
Shear modulus $G_{\mathrm{VRH}}$ depends on bond direction: it
depends on the directional stiffness of bonds and is suppressed in
low symmetry crystals where slip systems
proliferate~\citep{deJong2015elastic}. Structural splits (cubic vs.\
low-spacegroup) move the input along this axis, and the GNN must
encode symmetry information.

\paragraph{Encoder-bottlenecked: log\_kvrh, perovskites, phonons.}
Bulk modulus $K_{\mathrm{VRH}}$ measures resistance to volumetric
compression and scales mostly with mean bond stiffness and
density---an aggregate, near-isotropic
quantity~\citep{deJong2015elastic}. A GNN can predict
$K$ accurately while collapsing both compositional and structural
detail into a low-dimensional density-stiffness summary, leaving
nothing for a probe to find. The MatBench perovskites task is
restricted by construction to ABX$_3$ compositions in perovskite-like
structures~\citep{castelli2012perovskite}: most of our split rules
either select almost the entire dataset (oxide: most perovskite
entries are oxides) or leave the residual manifold the GNN already
covers, so the ``OOD'' set is not really out of the training
distribution. The highest optical phonon frequency depends on reduced masses
and local force constants~\citep{petretto2018phonons}; with a $64$--$256$ dimensional
embedding optimized to predict a scalar frequency, much of the
mode resolving information is thrown away. Across all three, the
GNN's training objective and the OOD split axes are roughly
orthogonal.

\subsection{Compositional shift magnitude vs.\ probe recoverability}
\label{app:chemistry_magpie}

To test the alignment claim quantitatively, we measure compositional
shift magnitude in a representation that does not depend on the GNN.
For each setting we compute the Magpie compositional
fingerprint~\citep{ward2016magpie} for every structure, take the mean
fingerprint vector for the ID and OOD pools, and define the Magpie
shift as the $L_2$ distance between these means after per-feature
standardization on the ID pool. Magpie features are elemental
statistics (electronegativity, atomic radius, valence counts, etc.)
aggregated over composition; they make no use of structure or learned
representations, so the resulting distance reflects raw compositional
dissimilarity. Implementation notes are in
Appendix~\ref{app:magpie_implementation}.

\begin{figure}[h]
  \centering
  \includegraphics[width=\linewidth]{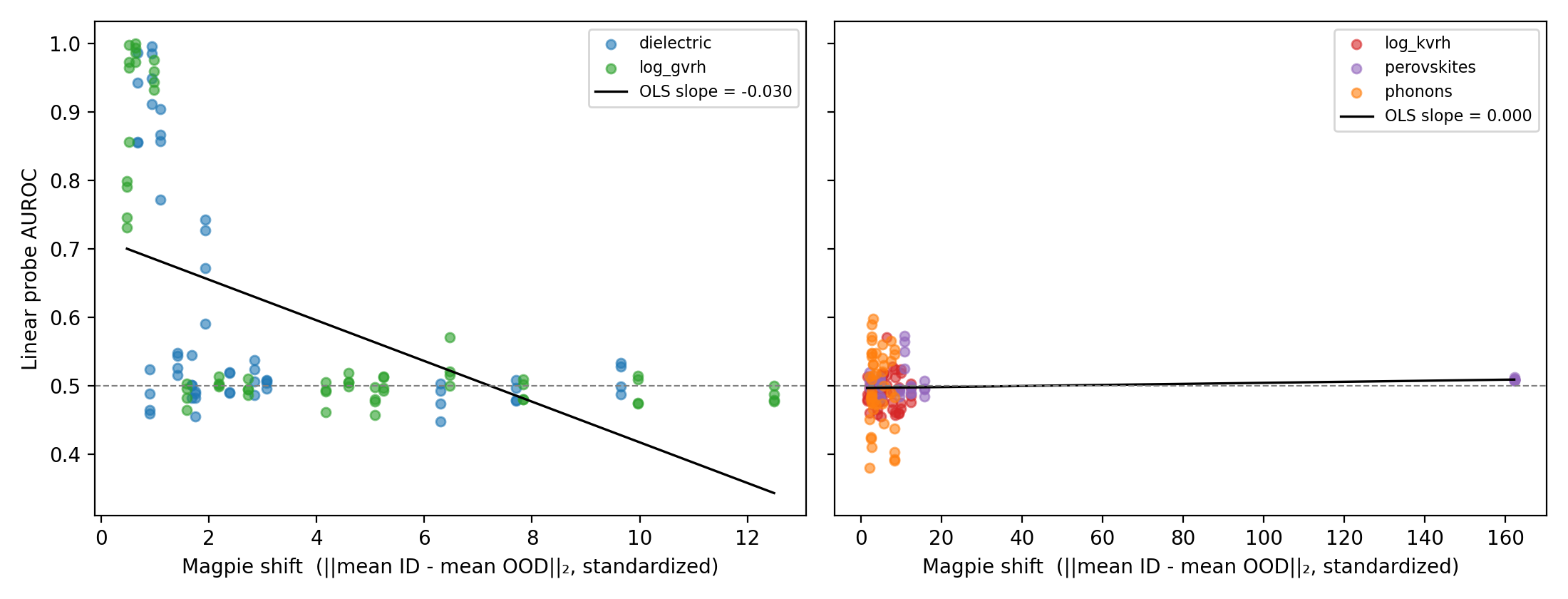}
  \caption{\textbf{Probe AUROC against composition-only shift
  magnitude.} The two regimes are distinguished by AUROC range, not
  slope: encoder-bottlenecked datasets are pinned at chance regardless
  of compositional shift (OLS slope $0.000$), while method-bottlenecked
  datasets reach probe AUROCs up to $1.0$ but the highest values
  cluster at low Magpie shift (OLS slope $-0.030$), showing that
  probe recoverability is governed by property relevant encoding
  rather than crude compositional distance.}
  \label{fig:magpie_vs_probe}
\end{figure}

Figure~\ref{fig:magpie_vs_probe} separates the two regimes by AUROC
range, not by slope. On the encoder-bottlenecked datasets (log\_kvrh,
perovskites, phonons), every probe AUROC sits within $\pm 0.02$ of
chance regardless of how different ID and OOD are in Magpie space; the
OLS slope is $0.000$. Compositional shifts that are large in raw
feature space do not produce probe-recoverable signal in the
embedding, the hallmark of an encoder bottleneck: information that
exists in the raw input does not survive the GNN's property-prediction
objective.

The method-bottlenecked datasets (dielectric, log\_gvrh) span probe
AUROCs from $0.5$ all the way to $1.0$, but this spread is not
correlated with Magpie shift: the OLS slope is $-0.030$, basically
flat or slightly negative. The highest AUROC settings cluster at low
Magpie shift---seven of the eight settings with probe AUROC $> 0.97$
are comp\_oxide or comp\_halide splits on log\_gvrh and dielectric, and
all four of these (dataset, split) pairs sit at Magpie shift $< 1.0$. These are the splits where most of the source
dataset is the ID class (oxides dominate the source pool), so the
mean composition difference is small in raw feature space, yet the
GNN encodes the split rule cleanly because oxide-vs.-non-oxide drives
the property's electronic structure. The contrast between the two
regimes is therefore not that one tracks Magpie shift and the other
doesn't---neither does in the linear slope sense. The contrast is
that method-bottlenecked embeddings can register the split rule when
the property physics demands it, while encoder-bottlenecked embeddings
cannot register it at all.

\subsection{Per-split breakdown}
\label{app:chemistry_per_split}

The dataset-level summary hides a lot of per-split structure. Two
patterns recur. First, splits that isolate a property-relevant composition class
(comp\_oxide and comp\_halide in particular) produce the highest probe
AUROCs on dielectric and log\_gvrh, often above $0.9$. High AUROC here
reflects a split rule that the GNN encodes cleanly because the property
depends on it, not element rarity: oxides are the majority of the
source pool. By contrast, comp\_intermetallic stays near chance
($\leq 0.53$). Second,
high spacegroup splits (sg $\geq 195$) are the weakest probe targets
even on method-bottlenecked datasets: high symmetry structures sit
near the centre of the training distribution rather than at its
edges, so the split rule picks out a denser region of the training
manifold rather than a genuinely shifted one. In several settings the
high-spacegroup ``OOD'' pool is closer to the ID mean in Magpie space
than a random ID subsample. Rule-based splits are not automatically
OOD; whether they are depends on the property and the data.

A practical takeaway for benchmark design: per-split AUROC should not
be averaged uniformly across split axes when the axes have different
a priori distances from the training distribution. We keep uniform
averaging for headline numbers in \S\ref{sec:eval_uniform} for
comparability with prior benchmarks, but caution that gains driven by
oxide/halide settings on dielectric should not be read as
gains on materials OOD detection broadly.

\subsection{Magpie analysis: implementation}
\label{app:magpie_implementation}

We use the Magpie elemental statistics implementation in
matminer~\citep{ward2018matminer} via the ElementProperty featurizer
with the magpie preset. For each composition we compute the minimum, maximum, range, mean,
average deviation, and mode of $22$ elemental properties, yielding a
$132$-dimensional fingerprint. Per-feature
means and standard deviations are computed on the ID pool only and
used to standardize both ID and OOD fingerprints. Setting-level shift
magnitudes are computed on the same balanced subsamples used for
AUROC evaluation, so sample size effects are controlled.

\subsection{Summary}

The bottleneck partition is not arbitrary. Method-bottlenecked
datasets have at least one split axis that is property relevant,
forcing the GNN to encode OOD defining variation; encoder-bottlenecked
datasets do not, and the GNN's compression throws away what little
OOD relevant signal exists. This sharpens
\S\ref{sec:eval_discussion}: the call for representation learning
that preserves OOD relevant structure is specifically a call to keep
variation along axes the property prediction objective treats as
nuisance, which the physics determines per dataset.

\FloatBarrier

\section{Usage guide}
\label{app:usage}

The minimal interface is a function score(z\_train, z\_test) that
returns a numpy array. It takes raw training and test embeddings (normalise inside the detector if needed) and returns one score per test point (higher =
more ID-like, matching the convention in Appendix~\ref{app:algorithms}). Wrapping this into the MOOD evaluation harness:
\begin{verbatim}
from mood.evaluate import evaluate_detector

def my_detector(z_train, z_test):
    # ... compute scores ...
    return scores

results = evaluate_detector(
    score_fn=my_detector,
    embeddings_root="data/embeddings/",
    output="results/my_detector.csv",
)
print(results.groupby("dataset")["auroc"].mean())
\end{verbatim}
The harness handles per-setting I/O, AUROC computation, and
aggregation; total runtime on the full 264-setting matrix is under one
CPU-hour for a typical post-hoc detector. Extending the benchmark axes
(new dataset, architecture, or split rule) is documented in the
README.md of the code release.

\FloatBarrier

\section{Author statement, licensing, and limitations}
\label{app:author_statement}

\paragraph{Responsibility.}
The authors take responsibility in case of violation of rights or other
issues arising from the data, code, or pre-extracted embeddings released
with this benchmark. Source-data ownership remains with the MatBench
authors and the Materials Project; pre-extracted embeddings and
aggregated AUROC tables are derivative works released under CC-BY-4.0
and credit the original MatBench publication.

\paragraph{Intended uses.}
MOOD is intended as a diagnostic benchmark for post-hoc OOD detection
on graph-structured scientific data. The evaluation matrix is designed
to support claims about whether existing detectors transfer to this
domain (the answer is largely no). Researchers proposing
new detectors should use MOOD as one of several benchmarks, not the
sole target.

\paragraph{Maintenance and growth commitment.}
We commit to maintaining and expanding MOOD as an actively supported
benchmark rather than a one-off release. Concretely, we plan to (1)
keep the code, splits, and pre-extracted embeddings hosted at the
URLs above for at least five years from publication, (2) issue
patch releases for any bugs reported through the repository's issue
tracker, (3) accept community contributions of new detectors,
architectures, and split rules through pull requests, and (4)
expand benchmark coverage in a v2 release that adds the missing
items called out in the limitations: AUPR-In / AUPR-Out across all
17 detectors, additional encoder architectures
(Matformer, MACE), and additional MatBench tasks. We will
version every release (v1, v2, $\ldots$) so prior results remain
reproducible against the snapshot they were computed on. We
encourage the community to propose new tasks, detectors, and split
axes; the README documents the contribution interface.

\paragraph{Licensing.}
\begin{itemize}[noitemsep,topsep=0pt]
    \item \textbf{Source code.} Released under the MIT License.
    \item \textbf{Embeddings and aggregated AUROC tables.} Released
        under CC-BY-4.0.
    \item \textbf{Source MatBench data.} Not redistributed; remains under
        the original Materials Project licensing terms.
\end{itemize}

\paragraph{Scope and limitations.}
\begin{itemize}[noitemsep,topsep=0pt]
    \item \textbf{Coverage.} 5 MatBench tasks and 4 architectures cover
        a portion of crystal-property prediction but do not cover
        molecular property prediction, force/energy prediction, or
        additional encoder architectures such as Matformer or MACE.
        Related benchmarks study OOD \emph{generalization} for molecules
        (DrugOOD~\citep{ji2023drugood}, BOOM~\citep{antoniuk2025boom}) and for geometric deep learning across scientific domains
        (GeSS~\citep{zou2024gess}), but none evaluates post-hoc detection.
    \item \textbf{Seed-dependent encoder collapse on phonons.} Seven of
        the $13$ CGCNN/phonons settings collapse to constant embeddings
        under seed 42 only (variance below $10^{-10}$); all seven train
        normally under seeds 123 and 456. This is a property of the
        single seed, not of CGCNN: we retain these settings and average
        over the two valid seeds (\S\ref{app:robustness}), so no setting
        is excluded and per-architecture comparisons on phonons use all
        four architectures.
    \item \textbf{Supervised probe regularisation mismatch.} The
        linear probe uses sklearn's C=1.0 (L2 strength = 1.0) and the
        MLP probe uses alpha=1.0 (matched). We checked empirically
        that the original sklearn-default $\alpha = 10^{-4}$ MLP and
        the $\alpha = 1.0$ MLP differ by $\leq 0.007$ mean AUROC on
        every dataset; the bottleneck classifications do not change.
        Both configurations are released for reproducibility.
    \item \textbf{FPR95 and AUPR coverage.} We report FPR95 for 14 of the 15 unsupervised detectors (all detectors from Table~\ref{tab:per_detector_summary} except Flow, whose FPR95 is stored separately and is deferred to v2 of this benchmark). AUPR-In / AUPR-Out across all 17 detectors require a cluster recompute and are deferred to the next benchmark version. Because the benchmark's reported AUROCs are computed on class-balanced ID/OOD test sets (subsampled to $\min(|\mathrm{ID}|, |\mathrm{OOD}|)$ per setting; \S\ref{sec:design}), AUPR-In and AUPR-Out under that balanced-sampling protocol are expected to track AUROC closely; deployment realism AUPR on the natural unbalanced test sets is a more informative companion measurement and is also deferred to v2.
    
    \item \textbf{Bottleneck threshold choice.} Our dataset
        classification uses a 0.55 probe AUROC threshold (five points
        above chance) and a 0.05 probe gap threshold. These are
        conservative cutoffs chosen to sit between the two regimes our
        datasets fall into: method-bottlenecked datasets reach mean
        probe AUROC 0.599--0.600, while encoder-bottlenecked datasets
        stay within 0.006 of chance. Under the mean-detector probe gap, no dataset sits near either threshold.
        Measured
        against the single best detector (Cosine), both method-bottlenecked datasets fall below
        the 0.05 gap cutoff and would be labeled Mixed rather than Method; Mixed still means the
        probe beats the best detector, by less than 0.05. The encoder-bottleneck labels depend
        on probe AUROC alone and are unaffected. We report both definitions in
        Table~\ref{tab:bottleneck_summary}. On a benchmark where datasets cluster differently,
        the thresholds may need to be re-anchored to the observed
        bimodality.
\end{itemize}

\paragraph{Data and code availability.} All artifacts for MOOD are
publicly released:
\begin{itemize}[noitemsep,topsep=0pt]
    \item \textbf{Code:}
        \url{https://github.com/vertaix/mood-benchmark}
    \item \textbf{Pre-extracted GNN embeddings, OOD splits, and
        detector outputs,} including the aggregated per-setting results
        (per\_seed\_cells.csv, perm\_per\_seed.csv) used by reproduce.py:
        \url{https://huggingface.co/datasets/shirlyu/mood-benchmark}
        (CC-BY-4.0).
    \item \textbf{Croissant metadata} (with Responsible AI fields):
        included in the OpenReview submission and on the HuggingFace
        dataset page.
\end{itemize}
Source MatBench data is not redistributed; users obtain it directly
from the Materials Project.

\end{document}